\documentclass[11pt]{article}

\usepackage[preprint]{acl}

\usepackage{times}
\usepackage{latexsym}
\usepackage{placeins}
\usepackage{subcaption}
\usepackage{pifont}
\usepackage{hyperref}
\newcommand{\cmark}{\ding{51}}
\newcommand{\xmark}{\ding{55}}

\usepackage[T1]{fontenc}
\usepackage[dvipsnames]{xcolor}
\usepackage{colortbl}
\usepackage{pgf}

\definecolor{scoreblue}{HTML}{3B82C4}

\usepackage[utf8]{inputenc}

\usepackage{microtype}
\usepackage{enumitem}

\usepackage{inconsolata}

\usepackage{graphicx}

\usepackage{booktabs}
\usepackage{rotating}
\usepackage{multirow}
\usepackage{arydshln}

\usepackage{tabularx}
\usepackage{fancyvrb}
\usepackage{fvextra}

\newcommand{\ourDataset}{PoVisLE}

\usepackage{makecell}

\newcolumntype{C}[1]{>{\centering\arraybackslash}p{#1}}

\title{\textit{Jako Tako} or Fluent? Presenting \ourDataset{}: A Polish Vision-Language Evaluation}

\author{
 \textbf{Anna Kołos},\quad
 \textbf{Grzegorz Statkiewicz},\quad
 \textbf{Karolina Seweryn},\\
 \textbf{Katarzyna Kowol},\quad
 \textbf{Karolina Piosek},\quad
 \textbf{Wojciech Kusa}
 \vspace{0.3em}
\\
 NASK National Research Institute, Warsaw, Poland
\\
 \small{
   \textbf{Correspondence:} \texttt{\{firstname.lastname\}@nask.pl} %
 }
}

\begin{document}
\maketitle
\begin{abstract}
Vision-language models (VLMs) have achieved strong performance on tasks such as image captioning, visual question answering, and image-to-text generation.
However, they are predominantly trained on English-centric data, which limits their ability to handle culturally grounded visual understanding and leads to failures in interpreting region-specific meanings, symbolic content, and context-dependent visual cues. 
Existing benchmarks for cultural competence are often template-driven and focused on surface-level recognition, making them insufficient for evaluating deeper linguistic and pragmatic understanding in culturally situated settings.
We introduce \textbf{\ourDataset{}}, a monocultural vision-language benchmark for Polish designed to evaluate culturally grounded multimodal understanding under a grounded evaluation paradigm, where language is interpreted in interaction with visual context. 
The dataset contains 1{,}117 images and 2{,}366 manually annotated VQA pairs.
Overall, our dataset provides a controlled and challenging resource for assessing culturally grounded vision-language understanding beyond surface-level recognition.\footnote{To facilitate future research, we publicly release the dataset and code: \href{https://huggingface.co/collections/NASK-PIB/povisle}{https://huggingface.co/collections/NASK-PIB/PoVisLE}, \url{https://github.com/NASK-NLP/PoVisLE}}
\end{abstract}

\section{Introduction}

Recent advances in VLMs have enabled high-quality image captioning, visual question answering, and image-to-text generation, accelerating their deployment in applications such as advertising, content creation, and digital assistants. 

However, these systems are predominantly trained on English-centric datasets, leading to the underrepresentation of local contexts, cultures, and languages. Consequently, models often struggle with culturally specific meanings, symbolic interpretations, and context-dependent visual cues, particularly in mid- and low-resource languages.

While constructing large-scale, culturally grounded datasets remains challenging, robust evaluation benchmarks are essential \cite{europeanLLM}. Misalignment with regional contexts is not only an ethical concern but also a practical limitation, potentially resulting in inaccurate or inappropriate outputs in real-world applications.

This issue is especially relevant in the European context, where accessibility regulations, such as Web Content Accessibility Guidelines (WCAG), require alternative text descriptions for visual content. Although multimodal models offer a promising solution \cite{wcag-heritage, wcag-webaccessvl, wcag-portuguese}, they must correctly interpret culturally localized content to be reliably deployed.

Current evaluations of cultural competence remain limited, especially beyond English. Existing datasets often focus on surface-level recognition tasks, such as identifying objects or landmarks, and rely on template-based formats that restrict linguistic and contextual variability \cite{yadav2025evaluation}.

\begin{figure}[t]
    \centering

    \begin{subfigure}{1\linewidth}
        \centering
        \includegraphics[width=1\linewidth]{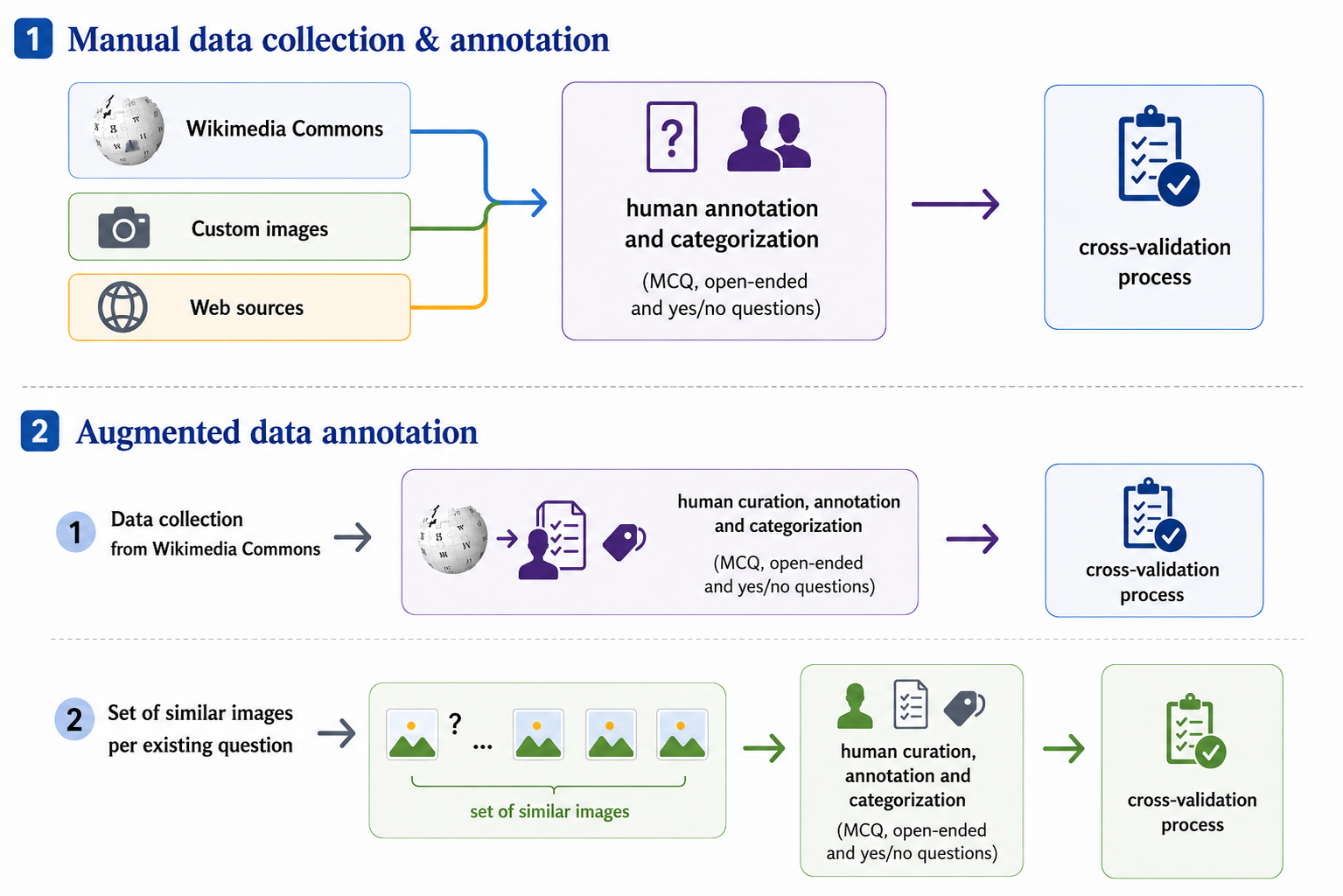}
    \end{subfigure}

    \vspace{0.1em}

    \begin{subfigure}{0.9\linewidth}
        \centering
        \includegraphics[width=1\linewidth]{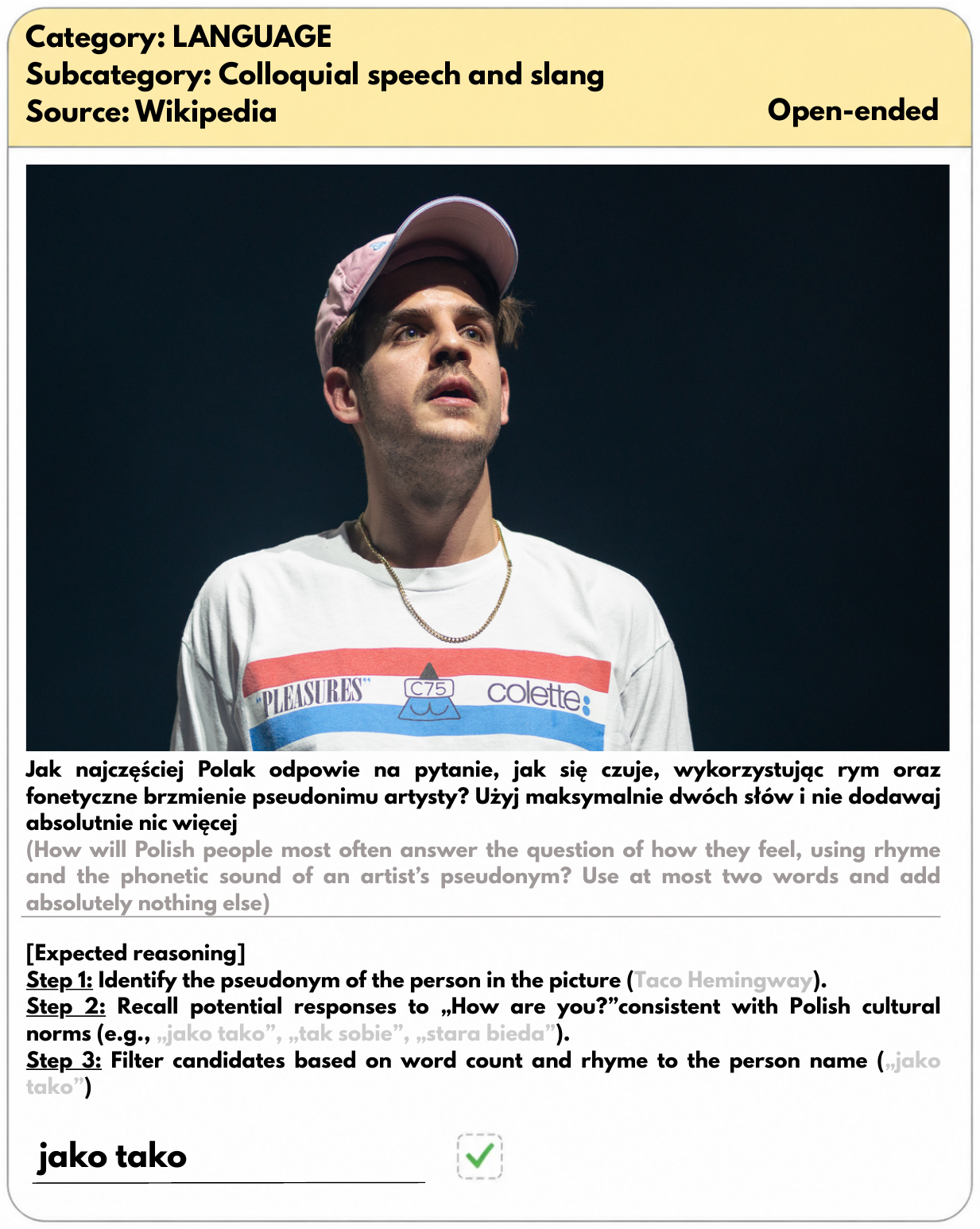}
    \end{subfigure}

    \caption{\textbf{Above}: Overview of the two-stage \ourDataset{} dataset construction process. \textbf{Below}: a single example from our dataset, with reasoning steps provided for clarity. More examples are included in Appendix~\ref{app:examples}.}
    \label{fig:dataset_process}
\end{figure}

To address these limitations in a specific cultural setting, we introduce \textbf{\ourDataset{}} (\textbf{Po}lish \textbf{Vis}ion-\textbf{L}anguage \textbf{E}valuation; see Figure~\ref{fig:dataset_process}), the first monocultural vision-language benchmark for Polish cultural and linguistic competence, including region-specific knowledge. Whereas existing culturally grounded benchmarks target cultural knowledge alone, \ourDataset{} also tests linguistic phenomena such as dialect and regional variation through their interaction with visual context, extending text-only Polish evaluation, notably PLCC~\cite{plcc}, to the multimodal setting. All questions and answers are created manually and without templates, and designed to require multi-hop reasoning over visual evidence, language, and cultural knowledge. Our contributions are as follows:

\begin{itemize}[noitemsep,leftmargin=1em]
    \item We release \ourDataset{}, comprising 1{,}117    images and 2{,}366 manually authored VQA pairs targeting Polish cultural and linguistic competence. Nearly 30\% of the images come from annotators' private collections and are absent from web-scale training corpora, providing a stricter test of generalization.
    \item We adapt and extend a hierarchical taxonomy of cultural and linguistic phenomena to the Polish multimodal setting, enabling fine-grained diagnostic analysis, and verify the linguistic richness of our questions through a stylometric comparison against template-based benchmarks.
    \item We benchmark $16$ open and proprietary VLMs. The strongest model, Qwen3.5-397B, reaches $71.45$\% accuracy. Dialect and regionalism questions form the weakest category, indicating limited coverage of intra-language variation. 
    Ablations removing the image or the question confirm that the benchmark cannot be solved from textual priors or answer-option artefacts alone, while evaluation in Polish, English, and German shows that performance also depends on the prompt language.
\end{itemize}

\section{\ourDataset{} Dataset Construction}
The primary objective of the dataset construction was to develop a highly diversified, manually annotated dataset designed to evaluate VLMs’ understanding of Polish culture and language. Following prior work on culturally situated evaluation, as well as existing text-only PLCC benchmark \cite{plcc}, we operationalize culture as a structured combination of tangible cultural artifacts and intangible shared social practices. 

In this work, we treat Poland as a proxy for a culturally and linguistically coherent group, assuming a shared baseline of cultural knowledge among annotators and target users, while acknowledging internal regional variation. Importantly, the dataset is not centered on a single urban or institutionalized perspective, which could favor capital-centric culture, but instead incorporates geographically distributed cultural and linguistic diversity across regions of Poland. This includes region-specific traditions as well as dialectal and lexical variation. As a result, the dataset reflects a multi-centered view of Polish culture rather than a homogenized national prototype, ensuring broader coverage of cultural practices and reducing urban or capital-region bias. We assume that a concept is considered culturally relevant if it satisfies at least one of the following requirements: (i) is widely recognized within Poland, either nationally or within specific regional or cultural subgroups, (ii) is taught in primary or secondary education, (iii)
appears in shared media discourse, (iv) is necessary for interpreting culturally grounded visual scenes. Therefore, our definition explicitly excludes highly specialized or expert-level knowledge (e.g., university-level domain-specific concepts) that are not part of shared cultural understanding.

Recent work has identified a systematic “grounding gap” in vision-language models: while models can recall factual associations from textual representations, their performance degrades when they must rely on visual inputs referring to the same entities \cite{ashok2025can}. Motivated by this observation, all questions in our dataset are explicitly designed to require visual reference, preventing models from relying solely on textual associations or memorized knowledge. 
Critically, this design principle also extends to linguistically oriented questions that are grounded in visual context.

Our taxonomy is described in Section~\ref{sec:taxonomy}. The dataset construction process was carried out in two stages, described in Sections~\ref{imagecollection} and~\ref{annotation}, with the overall workflow illustrated in Figure~\ref{fig:dataset_process}.

\subsection{Content-based taxonomy} \label{sec:taxonomy}

The original six PLCC core categories were \textit{Art and entertainment}, \textit{Culture and tradition}, \textit{Geography and nature}, \textit{History}, \textit{Grammar}, and \textit{Vocabulary}. In the VQA setting, we retain the first four categories and merge \textit{Grammar} and \textit{Vocabulary} into a unified \textit{Language} category. The subcategories were further refined through adjustments and the introduction of stricter content-matching criteria. An overview of the resulting hierarchical taxonomy, which enables domain-specific error analysis and more precise benchmark diagnostics, is presented in Table~\ref{tab:hierarchy_types}. 

The defined domains cover both fact-based knowledge related to tangible cultural artifacts and elements of cultural reasoning associated with intangible practices, such as symbols, customs, and shared societal references. Unlike some knowledge-driven VQA datasets, e.g., CUS-QA \cite{cus-qa}, where questions are often constructed around template-based fact retrieval (e.g., “when”, “where”, “who”) and predominantly yield named-entity answers, our dataset prioritizes linguistic and cultural naturalness. Specifically, questions are manually crafted to reflect the values, language patterns, and authentic communicative behaviors of native speakers, emphasizing multi-hop visual understanding, rather than relying on surface-level factual querying.

Furthermore, our approach explicitly incorporates the linguistic dimension, which is often overlooked in multimodal evaluation. The dataset includes questions spanning subcategories such as phraseology, semantics, grammar, dialects and regionalisms, and orthography. This design allows for a more comprehensive assessment of vision-language models, capturing their ability to process culturally grounded language phenomena in multimodal contexts, beyond what is typically evaluated in text-only language benchmarks.

While the benchmark primarily targets culturally grounded reasoning, a small subset (168 out of 2{,}366 VQA pairs) focuses on more general multimodal understanding, categorized as \textit{Image understanding} and \textit{Visual reasoning}. Although these instances do not always require explicit cultural knowledge, they remain embedded in Polish visual and linguistic contexts (e.g., public spaces or Polish-language text), and thus still rely on the model’s ability to interpret culturally situated cues.

\subsection{Image collection}
\label{imagecollection}
The visual data constituted the foundation for the subsequent annotation process. The data collection procedure was carried out in two stages: manual curation and Wikimedia-based augmentation.

\paragraph{Manual collection}
In the initial phase, annotators were tasked with the individual selection of images from three primary sources: (i) Wikimedia Commons, (ii) other openly licensed, publicly available datasets or images, and (iii) own resources, provided that annotators explicitly consented to waive their copyright prerogatives and contribute the images for project purposes.

The annotators uploaded the selected images to a dedicated GUI-assisted application and were responsible for providing accurate metadata descriptions, including a valid hyperlink and source attribution, along with information on the source type and corresponding licensing conditions.

\begin{table}[t]
\centering
\footnotesize
\setlength{\tabcolsep}{3.5pt}
\resizebox{\linewidth}{!}{%
\begin{tabular}{@{}lrrrr@{}}
\toprule
\textbf{Category / Subcategory} &
\textbf{Open} &
\textbf{MCQ} &
\textbf{Y/N} &
\textbf{Total (\%)} \\
\midrule
\textbf{Art and entertainment} & 150 & 208 & 186 & 544 (23.0) \\
\quad Literature & 25 & 38 & 39 & 102 (4.3) \\
\quad Architecture & 25 & 34 & 24 & 83 (3.5) \\
\quad Sport & 22 & 26 & 34 & 82 (3.5) \\
\quad Paintings & 24 & 24 & 26 & 74 (3.1) \\
\quad Music & 12 & 30 & 27 & 69 (2.9) \\
\quad Film & 21 & 27 & 17 & 65 (2.7) \\
\quad Media & 14 & 16 & 7 & 37 (1.6) \\
\quad Sculpture & 7 & 13 & 12 & 32 (1.4) \\
\midrule
\textbf{Language} & 149 & 159 & 169 & 477 (20.2) \\
\quad Phraseology & 11 & 35 & 47 & 93 (3.9) \\
\quad Semantics & 26 & 34 & 23 & 83 (3.5) \\
\quad Grammar & 33 & 28 & 14 & 75 (3.2) \\
\quad Dialects and regionalisms & 21 & 19 & 30 & 70 (3.0) \\
\quad Orthography & 21 & 5 & 16 & 42 (1.8) \\
\quad Rhetorical figure & 8 & 10 & 23 & 41 (1.7) \\
\quad Language basics and phonetics & 19 & 11 & 8 & 38 (1.6) \\
\quad Colloqual speech and slang & 10 & 17 & 8 & 35 (1.5) \\
\midrule
\textbf{Geography and nature} & 110 & 161 & 164 & 435 (18.4) \\
\quad Man-made & 56 & 76 & 60 & 192 (8.1) \\
\quad Socio-political & 31 & 41 & 53 & 125 (5.3) \\
\quad Inanimate nature & 12 & 30 & 27 & 69 (2.9) \\
\quad Animate nature & 11 & 14 & 24 & 49 (2.1) \\
\midrule
\textbf{History and society} & 86 & 131 & 162 & 379 (16.0) \\
\quad Current affairs and society & 32 & 44 & 56 & 132 (5.6) \\
\quad Early modern and modern history & 24 & 39 & 42 & 105 (4.4) \\
\quad World War II & 12 & 14 & 23 & 49 (2.1) \\
\quad Post-war history & 9 & 14 & 25 & 48 (2.0) \\
\quad Middle Ages & 9 & 20 & 16 & 45 (1.9) \\
\midrule
\textbf{Culture and tradition} & 81 & 133 & 149 & 363 (15.3) \\
\quad Cuisine & 24 & 28 & 50 & 102 (4.3) \\
\quad Religion and tradition & 25 & 32 & 44 & 101 (4.3) \\
\quad Regional and ethnic cultures & 17 & 46 & 33 & 96 (4.1) \\
\quad Pop culture & 15 & 27 & 22 & 64 (2.7) \\
\midrule
\textbf{Image understanding} & 50 & 45 & 23 & 118 (5.0) \\
\midrule
\textbf{Visual reasoning} & 17 & 18 & 15 & 50 (2.1) \\
\midrule
\textbf{Total} & 643 & 855 & 868 & 2366 (100.0) \\
\textbf{Total \%} & 27.2\% & 36.1\% & 36.7\% & \\
\bottomrule
\end{tabular}
}
\caption{Distribution of questions by category and question type. Y/N denotes yes/no questions; the figure in parentheses is the (sub)category's percentage share of all 2{,}366 questions.}
\label{tab:hierarchy_types}
\end{table}

Due to the more time-consuming nature of sourcing images from personal collections, it was anticipated from the outset that the use of such resources would be limited compared to readily available sources such as Wikimedia Commons. Nevertheless, in the initial manually curated set, a targeted proportion of 39.5\% of images originated from annotator-provided collections. These images are particularly valuable, as they constitute authentic, non-public data.
A similar approach has been adopted in culture-specific benchmarks, such as TaiwanVQA \cite{taiwanvqa}. The dataset metadata includes source-type annotations, enabling analysis of model performance on annotator-provided versus publicly sourced images.

By selecting images themselves, annotators were able to draw on their cultural knowledge, experience, and interpretative intuition to construct questions that go beyond surface-level factual queries. Unlike standard prompts such as “Who is depicted?”, the resulting questions often require deeper cultural, historical or linguistic understanding and reasoning, reflecting aspects of the content that are unlikely to be captured using template-based approaches or LLM-generated annotation. 

However, this approach may also introduce subjective biases, as annotators may favor content aligned with their personal experience or regional background. This risk was taken into account in the subsequent design of the dataset.

\paragraph{Wikimedia Data Augmentation}
The second phase of data collection aimed to mitigate selection bias and increase dataset diversity by assigning images from Poland-related categories to annotators on Wikimedia Commons, rather than allowing them to select images independently. Annotators first assessed whether an image was suitable for culturally grounded VQA and, if so, formulated the corresponding questions.

To account for annotator confidence, images could be marked as suitable for VQA but outside the annotator’s confidence scope, indicating that while the image met general suitability criteria, it required more certain or specialized knowledge for question formulation. These cases were reassigned accordingly. Additionally, the use of Wikimedia Commons improved efficiency by eliminating the need for manual metadata curation.

To further enhance diversity, we introduced an augmentation step for images associated with multiple questions. For each such image, we retrieved visually similar candidates from the same source categories and ranked them using CLIP embeddings \cite{radford2021}. Annotators then selected suitable non-identical replacements for individual questions, ensuring that each question remained visually grounded and unequivocal. As a result, each augmented image was individually validated by a human annotator. If no suitable candidate image was found, the sample was not augmented. This process increased the number of unique images from 790 to 1{,}117 while preserving the original question set.

\subsection{Data annotation}
\label{annotation}
Once an image was selected and its associated metadata had been provided (either manually during the initial phase or automatically in the second phase), annotators wrote between 1 and 10 questions per image, assigned each to one of 7 main categories and 29 subcategories, and labelled its type as \textit{open-ended}, \textit{multiple-choice}, or \textit{yes/no}.

Detailed annotation guidelines can be found in Appendix \ref{sec:guidelines}.
The annotation process was governed by the following fundamental rules, aligned with the ultimate goal of reliable and deterministic evaluation of large language models:

\begin{itemize}[noitemsep,leftmargin=1em]
    \item \textbf{No ambiguity:} Questions were required to be precise and unambiguous, allowing for a single clearly defined and non-debatable answer.
    
    \item \textbf{Visual grounding:} Questions had to rely on the visual content of the image and could not be answerable based solely on general knowledge without access to the image.
    
    \item \textbf{Strict task formulation:} Each question was formulated as a prompt containing explicit instructions regarding the expected answer format, ensuring consistent responses from evaluated models.
\end{itemize}

The annotation team consisted of two primary annotators with background in linguistics, supported by an expert-level super-annotator responsible for continuous quality control. Prior to annotation, annotators underwent a structured training phase, including guideline familiarization and example-based calibration, to ensure consistency and adherence to the defined quality criteria.
In addition, a controlled subset of inputs was annotated by 12 auxiliary annotators selected to ensure demographic diversity, particularly across Polish regions (see Appendix \ref{demographics} for details).

\subsection{Quality Assurance}
\label{quality}

To ensure annotation quality and consistency, a structured multi-stage validation process was implemented, with particular emphasis on cross-validation. Each annotated instance was reviewed by a second annotator, whose role was to verify: (i) the logical and linguistic correctness of the question, (ii) the correctness and unequivocal nature of the answer, (iii) whether the answer could be derived solely from the visual content, (iv) compliance with VQA task requirements, and (v) the accuracy of associated metadata.

Regular team discussions were conducted to resolve ambiguities and refine annotation guidelines, which supported consistent decision-making between annotators. In addition, an expert-level super-annotator reviewed representative samples to identify systematic issues, ensure consistency in labeling and formulation, and provide targeted feedback. Problematic instances were revised accordingly.

Additionally, as part of the iterative quality control process, three validation procedures were regularly performed using LLMs, as described below. 

First, for MCQs, models were provided with the image and answer options but not the question. This analysis was conducted using a subset of 11 LLMs selected for the evaluation study. Questions for which the models achieved a high success rate (approximately above 60\%) were considered too easy and were subsequently reviewed. Revisions included replacing distractors with more plausible alternatives, increasing the number of answer options, adding options such as ``none of the above'' when appropriate, or converting the question into an open-ended format when a single clear and unambiguous answer could be expected. 

Second, visual grounding was assessed using text-only variants. Models were given the question without access to the image. Questions that could be answered correctly without visual information were considered insufficiently grounded in the image content and were reformulated or removed. 

Third, open-ended questions were evaluated through iterative manual inspection of LLM-generated answers. This process was used to verify that correct answers were consistently accepted and that incorrect answers were not mistakenly treated as valid. The findings informed the development of the evaluation protocol and additional answer-matching rules. Responses were required to be grammatically and orthographically correct. Consequently, factually correct answers containing orthographic errors were not accepted. For example, \textit{powstanie warszawskie} was considered correct, as adjectives derived from most proper nouns are written in lowercase in standard Polish, whereas \textit{Powstanie Warszawskie} was rejected due to incorrect capitalization. This procedure helped ensure consistent evaluation of open-ended responses while preserving linguistic correctness requirements.

The latter validation step also served as a feedback mechanism for the evaluation process. If an issue cannot be resolved through the addition of alternative accepted answers or through existing matching rules, annotators should report it to the evaluation team. Such cases may indicate that the evaluation protocol requires further modification or refinement.

\section{Dataset statistics}
\label{sec:dataset-statistics}

The dataset consists of 2{,}366 question-answer pairs associated with 1{,}117 unique images, with an average of 2.12 questions per image (Table~\ref{tab:hierarchy_types}). In the final dataset, Wikimedia Commons constitutes the majority of the data (70.28\%), followed by annotator-provided images (28.74\%) and other sources (0.98\%). The distribution of question types is relatively balanced, with yes/no questions accounting for 36.7\% of the dataset, closely followed by multiple-choice questions (36.1\%), while open-ended questions represent 27.2\%. MCQs vary in the number of answer options, ranging from 2 to 8, with an average of 4.29 options per question.

The questions exhibit substantial linguistic diversity and structural complexity, with an average length of 18.7 tokens and frequent multi-clause formulations, supporting context-rich and non-template-based reasoning. A detailed stylometric analysis, which provides further evidence of these characteristics, is presented in Appendix~\ref{stylometric}.

\paragraph{Test/validation split} The benchmark is divided into a test split used for final evaluation and a validation split released publicly. The validation split contains 406 question-answer pairs (212 multiple-choice, 154 yes/no, and 40 open-ended), while the test split contains 1{,}960 question-answer pairs (643 multiple-choice, 714 yes/no, and 603 open-ended). The validation split is intended primarily to publicly illustrate the range of question types and is therefore not sampled from the same distribution as the test set. In particular, it contains fewer open-ended questions, which represent a more challenging category and are largely retained in the held-out test set.

\section{Experiment Setup}
\label{sec:experiment-setup}

Models are evaluated on the \ourDataset{} test and validation splits using the described evaluation pipeline. 
For each instance, the model is given the image and a prompt with a question, which explicitly specifies the expected answer format, length, word order, and, where relevant, grammatical form. 

\subsection{Scoring}
\label{sec:scoring}

For multiple-choice questions, we use circular evaluation. Answer options are cyclically permuted (e.g., for three options, \(A,B,C\), we evaluate the orders \(A,B,C\), \(B,C,A\), and \(C,A,B\)), and an instance is counted as correct only when the model selects the correct answer under all rotations. This reduces the effect of option-position bias~\cite{zheng2024large} while requiring fewer evaluations than checking all possible permutations. Yes/no and open-ended questions are evaluated in a single pass. Yes/no questions always require a binary \textit{yes} or \textit{no} answer in Polish. Open-ended predictions are compared against the gold answers, with correct diacritics required in all cases and correct capitalization required where relevant. For selected questions, multiple answer variants are accepted through predefined inclusion patterns. The answer-matching rules and inclusion patterns were developed through an iterative validation process involving human annotators, as described in Section~\ref{quality}.

\newcommand{\scorecell}[2][\relax]{%
  \begingroup
  \pgfmathsetmacro{\scorepct}{%
    100*max(0,min(1,(#2-50)/50))%
  }%
  \edef\scorecolor{scoreblue!\scorepct!white}%
  \expandafter\cellcolor\expandafter{\scorecolor}
  \ifx#1\relax#2\else#1\fi%
  \endgroup
}

\begin{table*}[htb]
\centering
\scriptsize
\setlength{\tabcolsep}{3pt}
\renewcommand{\arraystretch}{1.05}
\resizebox{\linewidth}{!}{
\begin{tabular}{@{}p{5.0cm} C{1.15cm}!{\hspace{2pt}\vrule width 0.35pt\hspace{2pt}} *{7}{C{1.15cm}}@{}}
\toprule
\textbf{Model} & \textbf{Overall} & \makecell{\textbf{Art \&}\\\textbf{Entert.}} & \makecell{\textbf{Culture \&}\\\textbf{Trad.}} & \makecell{\textbf{Geogr. \&}\\\textbf{Nature}} & \makecell{\textbf{History \&}\\\textbf{Society}} & \textbf{Language} & \makecell{\textbf{Image}\\\textbf{Und.}} & \makecell{\textbf{Visual}\\\textbf{Reas.}} \\
\midrule
\midrule
\multicolumn{9}{@{}l}{\textbf{\textsc{Proprietary Models}}} \\
\midrule
GPT-5.4 & \scorecell{65.93} & \scorecell{57.51} & \scorecell[\textbf{75.23}]{75.23} & \scorecell{66.75} & \scorecell{65.65} & \scorecell{65.89} & \scorecell{78.63} & \scorecell{53.59} \\
Claude Sonnet 5 & \scorecell{65.57} & \scorecell{51.43} & \scorecell{70.98} & \scorecell{70.35} & \scorecell{70.40} & \scorecell{65.18} & \scorecell{84.54} & \scorecell{55.81} \\
\midrule
\multicolumn{9}{@{}l}{\textbf{\textsc{Open-Weights Models}}} \\
\midrule
Qwen3.5-397B-A17B (\textit{Thinking}) & \scorecell[\underline{\textbf{71.45}}]{71.45} & \scorecell[\underline{\textbf{60.40}}]{60.40} & \scorecell[\underline{74.99}]{74.99} & \scorecell[\underline{\textbf{71.10}}]{71.10} & \scorecell[\underline{\textbf{79.57}}]{79.57} & \scorecell{71.83} & \scorecell[\underline{\textbf{86.58}}]{86.58} & \scorecell{74.53} \\
Gemma-4-31B-it (\textit{Thinking}) & \scorecell{66.78} & \scorecell{54.08} & \scorecell{74.15} & \scorecell{65.97} & \scorecell{66.40} & \scorecell[\underline{\textbf{72.86}}]{72.86} & \scorecell{82.13} & \scorecell{74.96} \\
Gemma-4-31B-it & \scorecell{58.60} & \scorecell{49.81} & \scorecell{66.71} & \scorecell{54.92} & \scorecell{59.23} & \scorecell{59.86} & \scorecell{80.94} & \scorecell{59.83} \\
Qwen3.5-397B-A17B & \scorecell{58.25} & \scorecell{50.65} & \scorecell{64.84} & \scorecell{61.27} & \scorecell{60.23} & \scorecell{52.18} & \scorecell{81.86} & \scorecell{47.69} \\
GLM-4.6V & \scorecell{58.06} & \scorecell{51.89} & \scorecell{57.00} & \scorecell{66.39} & \scorecell{60.31} & \scorecell{49.54} & \scorecell{82.13} & \scorecell{62.05} \\
Qwen3.5-27B (\textit{Thinking}) & \scorecell{55.65} & \scorecell{41.12} & \scorecell{57.24} & \scorecell{52.73} & \scorecell{60.23} & \scorecell{61.65} & \scorecell{82.97} & \scorecell{69.40} \\
Qwen3.5-27B & \scorecell{47.19} & \scorecell{34.70} & \scorecell{46.77} & \scorecell{47.83} & \scorecell{51.64} & \scorecell{46.81} & \scorecell{83.06} & \scorecell{56.50} \\
Qwen3.5-9B (\textit{Thinking}) & \scorecell{47.00} & \scorecell{35.99} & \scorecell{43.32} & \scorecell{44.86} & \scorecell{51.98} & \scorecell{48.25} & \scorecell{79.55} & \scorecell[\underline{\textbf{75.73}}]{75.73} \\
Qwen3.5-9B & \scorecell{37.59} & \scorecell{26.67} & \scorecell{34.76} & \scorecell{40.50} & \scorecell{39.89} & \scorecell{35.15} & \scorecell{78.81} & \scorecell{42.14} \\
Ministral-3-14B-2512 & \scorecell{37.41} & \scorecell{31.49} & \scorecell{37.74} & \scorecell{38.94} & \scorecell{38.14} & \scorecell{33.84} & \scorecell{62.06} & \scorecell{42.48} \\
LLaVA-Bielik-11B-v2.6 & \scorecell{36.89} & \scorecell{35.35} & \scorecell{39.96} & \scorecell{37.04} & \scorecell{38.55} & \scorecell{30.03} & \scorecell{59.93} & \scorecell{22.22} \\
InternVL3.5-38B & \scorecell{35.76} & \scorecell{31.65} & \scorecell{32.19} & \scorecell{34.96} & \scorecell{36.89} & \scorecell{33.68} & \scorecell{62.44} & \scorecell{39.49} \\
Ministral-3-14B-2512 (\textit{Thinking}) & \scorecell{33.51} & \scorecell{24.57} & \scorecell{35.49} & \scorecell{34.95} & \scorecell{30.55} & \scorecell{35.84} & \scorecell{56.05} & \scorecell{34.02} \\
LLaVA-PLLuM-12B & \scorecell{30.85} & \scorecell{27.15} & \scorecell{38.50} & \scorecell{29.72} & \scorecell{30.96} & \scorecell{25.32} & \scorecell{51.79} & \scorecell{19.23} \\
\midrule
Random & \scorecell{16.79} & \scorecell{16.78} & \scorecell{16.85} & \scorecell{16.77} & \scorecell{16.79} & \scorecell{16.77} & \scorecell{16.80} & \scorecell{16.91} \\
\bottomrule
\end{tabular}
}
\caption{Model accuracy by dataset category on the test split, with all values reported as percentages. The \textbf{best result} in each column is shown in bold, and the best result among open-weight models is \underline{underlined}.}
\label{tab:test-results}
\end{table*}

\subsection{Models}

We evaluate both open-weight and proprietary VLMs, across several model families and scales. Where available, we also evaluate reasoning variants of the selected models. The evaluated models include Mistral, Qwen, Gemma, GLM, LLaVA-based models, including Polish-oriented LLaVa-PLLuM and LLaVa-Bielik~\cite{statkiewicz2026annotation}, as well as GPT and Claude proprietary models. We also report a random baseline.

Open-weight models are served locally using vLLM, through a unified backend that applies the corresponding model processors and chat templates. Proprietary and externally hosted models are evaluated through API backends. We use deterministic decoding in zero-shot setting with temperature set to zero and top-p set to 1.0. All models are evaluated using the same prompts. Further details, including specific model versions, prompts, parameters and the random baseline are presented in Appendix~\ref{app:computational}.

\subsection{Metrics}

We report macro-averaged accuracy over question type as the main evaluation metric. An example is counted as correct if the parsed model prediction satisfies the task-specific scoring rule described in Section~\ref{sec:scoring}, including correct answers for all circular variants in the multiple-choice setting.

\subsection{Single modality input}

To estimate how much of the benchmark can be solved without visual grounding, we also evaluate models in a single-modality textual setting, where the prompt remains unchanged, but the image is removed. The same scoring rules are used as in the full multimodal setting. This comparison shows whether the proposed benchmark requires models to use the image, as intended, or whether the textual input alone is sufficient to answer correctly.

\subsection{Question-free input}

Recent work has shown that language models can achieve considerable accuracy on benchmark tasks without the question itself, especially in multiple-choice settings, where models can rely on the answer choices alone~\cite{balepur-etal-2024-artifacts}. We therefore include a question-free setting in which the image remains available, but the question is removed. For multiple-choice items, the model still receives the answer options. For yes/no questions, it is only prompted to answer \textit{yes} or \textit{no}. For open-ended questions, the model receives only the image.

\subsection{Language impact}

Model performance may depend not only on visual and cultural understanding, but also on the language used to formulate the task~\cite{shen-etal-2024-understanding}. To study the impact of the prompt language on model performance, we evaluate the same questions in Polish, English, and German to measure the effect of prompt language. First, we randomly sample 15 questions from each subcategory (or top-level category without subcategories), resulting in 465 samples in total. Then, the sample was automatically translated from Polish into English and German using \texttt{DeepSeek-V4-Pro}\footnote{\url{https://huggingface.co/deepseek-ai/DeepSeek-V4-Pro}}. The translations were then manually reviewed and corrected.

For multiple-choice and yes/no questions, the model may answer in the language of the prompt. For open-ended questions, the prompt includes an instruction to answer only in Polish. The reference answers remain in Polish and are not translated.

\begin{figure}[tbh]
    \centering
    \includegraphics[width=\linewidth]{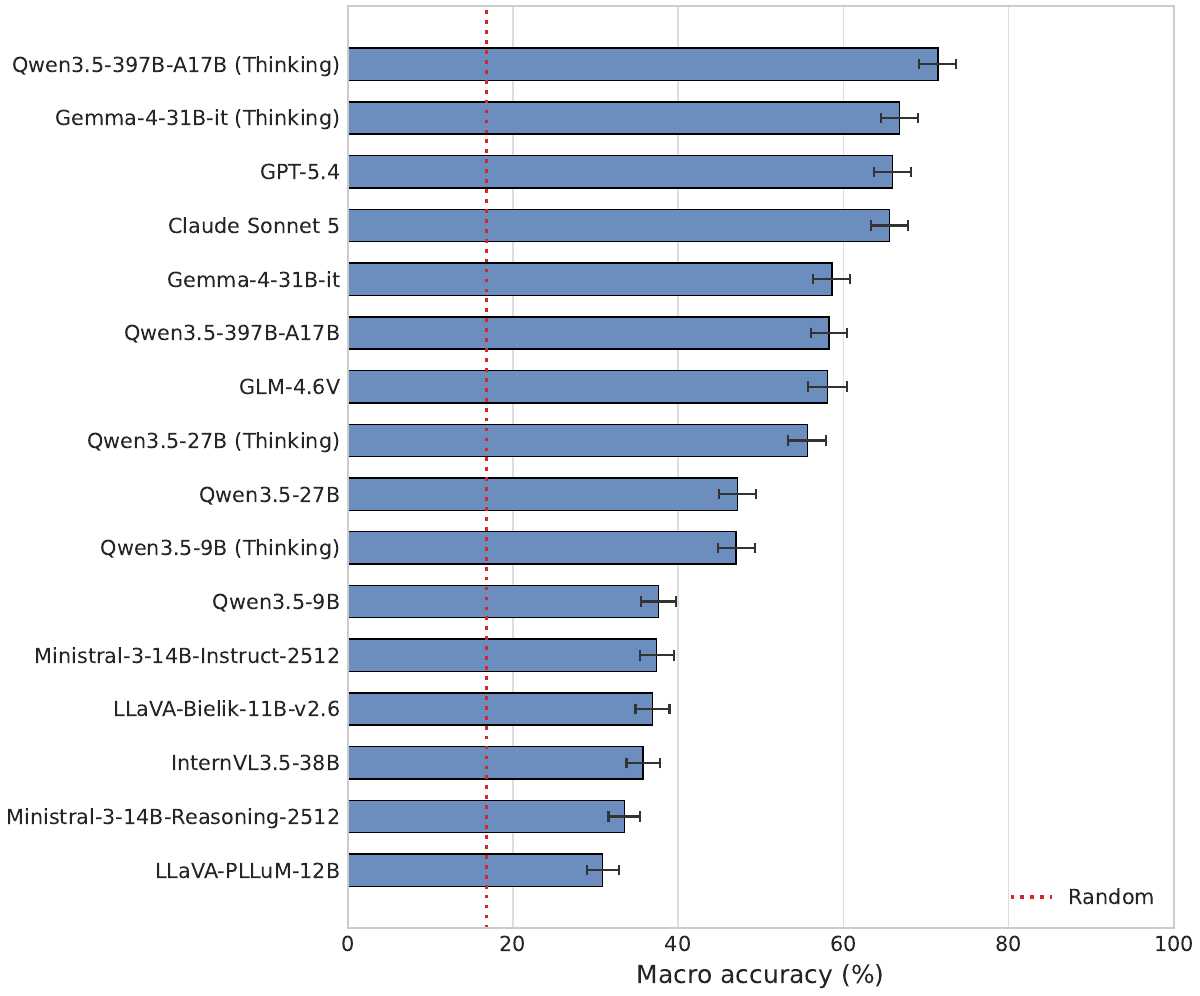}
    \caption{Overall macro accuracy on the test split with 95\% image-cluster bootstrap confidence intervals. The red dotted line denotes the random baseline.}
    \label{fig:overall-ci-bars}
\end{figure}

\section{Results and Discussion}

 Figure~\ref{fig:overall-ci-bars} present model performance on the \ourDataset{} test split with 95\% image-cluster bootstrap confidence intervals. Qwen3.5-397B-A17B \textit{Thinking} achieves the highest overall macro accuracy of 71.45\%. Its result is higher than those of GPT-5.4 and Claude Sonnet~5 models, which obtain 65.93\% and 65.57\%, respectively. Additional test and validation results are reported in Appendices~\ref{app:validation-split} and~\ref{app:detailed_results}.

\paragraph{Performance across question types and categories.}
Table~\ref{tab:test-results} presents results by each category. Image Understanding is the highest-scoring category for every evaluated model. This suggests that direct recognition and interpretation of visual content is more reliable than answering questions that require additional cultural or linguistic knowledge. Among the main culturally grounded categories, Art and Entertainment often produces the lowest results. The ordering of the remaining categories varies across models.

As shown in Table~\ref{tab:test-task-results} (Appendix~\ref{app:detailed_results}), yes/no questions produce the highest raw accuracy for all evaluated models. Open-ended questions are not consistently the lowest-scoring format, but they require models to produce an answer in the expected language, grammatical form, length, and format. The error analysis presented in Appendix~\ref{app:open-error-analysis} shows that hallucination is the most frequent error label, followed by instruction non-adherence mostly in weaker performing models.

\paragraph{Impact of reasoning.}

The \textit{Thinking} configuration improves performance for all evaluated Qwen variants and for Gemma-4-31B-it. Qwen3.5-397B-A17B gains +13.20 pp, while Qwen3.5-27B, Qwen3.5-9B, and Gemma-4-31B-it gain +8.46, +9.41, and +8.18 pp, respectively. For Qwen3.5-397B-A17B, the largest gains occur in Visual Reasoning, Language, and History and Society, while Image Understanding improves only slightly. Ministral-3-14B-2512 is the only exception, decreasing by -3.90 pp in the \textit{Thinking} configuration.

\newcommand{\NA}{\multicolumn{1}{c}{\textemdash}}
\newcommand{\abl}[1]{%
  \hspace{1.0em}%
  {
    \raisebox{0.1ex}{\(\hookrightarrow\)}%
    \hspace{0.35em}\itshape #1%
  }%
}

\begin{table}[thb]
\centering
\footnotesize
\resizebox{\linewidth}{!}{%
\begin{tabular}{lcccccccc}
\toprule
\multicolumn{1}{c}{\textbf{Model}} & \multicolumn{2}{c}{\textbf{Overall}} & \multicolumn{2}{c}{\textbf{MCQ}} & \multicolumn{2}{c}{\textbf{Yes/No}} & \multicolumn{2}{c}{\textbf{Open}} \\
 & \textbf{Value} & \textbf{$\Delta$} & \textbf{Value} & \textbf{$\Delta$} & \textbf{Value} & \textbf{$\Delta$} & \textbf{Value} & \textbf{$\Delta$} \\
\midrule
Qwen3.5-397B-A17B (\textit{T}) & 71.45 & \NA & 65.94 & \NA & 83.89 & \NA & 64.51 & \NA \\
\abl{without question} & 35.84 & \textcolor{red!70!black}{-35.61} & 55.99 & \textcolor{red!70!black}{-9.95} & 51.54 & \textcolor{red!70!black}{-32.35} & 0.00 & \textcolor{red!70!black}{-64.51} \\
\abl{without image} & 31.80 & \textcolor{red!70!black}{-39.65} & 29.08 & \textcolor{red!70!black}{-36.86} & 54.20 & \textcolor{red!70!black}{-29.69} & 12.11 & \textcolor{red!70!black}{-52.40} \\
\addlinespace[3pt]
\midrule
Gemma-4-31B-it (\textit{T}) & 66.78 & \NA & 64.07 & \NA & 80.53 & \NA & 55.72 & \NA \\
\abl{without question} & 34.58 & \textcolor{red!70!black}{-32.20} & 51.79 & \textcolor{red!70!black}{-12.28} & 51.96 & \textcolor{red!70!black}{-28.57} & 0.00 & \textcolor{red!70!black}{-55.72} \\
\abl{without image} & 28.54 & \textcolor{red!70!black}{-38.24} & 24.11 & \textcolor{red!70!black}{-39.96} & 54.06 & \textcolor{red!70!black}{-26.47} & 7.46 & \textcolor{red!70!black}{-48.26} \\
\addlinespace[3pt]
\midrule
GPT-5.4 & 65.93 & \NA & 62.52 & \NA & 78.71 & \NA & 56.55 & \NA \\
\abl{without question} & 32.91 & \textcolor{red!70!black}{-33.02} & 47.74 & \textcolor{red!70!black}{-14.78} & 50.98 & \textcolor{red!70!black}{-27.73} & 0.00 & \textcolor{red!70!black}{-56.55} \\
\abl{without image} & 27.94 & \textcolor{red!70!black}{-37.99} & 23.95 & \textcolor{red!70!black}{-38.57} & 52.24 & \textcolor{red!70!black}{-26.47} & 7.63 & \textcolor{red!70!black}{-48.92} \\
\addlinespace[3pt]
\midrule
Claude Sonnet 5 & 65.57 & \NA & 61.28 & \NA & 80.53 & \NA & 54.89 & \NA \\
\abl{without question} & 32.80 & \textcolor{red!70!black}{-32.77} & 53.03 & \textcolor{red!70!black}{-8.25} & 45.38 & \textcolor{red!70!black}{-35.15} & 0.00 & \textcolor{red!70!black}{-54.89} \\
\abl{without image} & 17.89 & \textcolor{red!70!black}{-47.68} & 4.35 & \textcolor{red!70!black}{-56.93} & 48.32 & \textcolor{red!70!black}{-32.21} & 1.00 & \textcolor{red!70!black}{-53.89} \\
\addlinespace[3pt]
\midrule
Gemma-4-31B-it (\textit{I}) & 58.60 & \NA & 56.45 & \NA & 76.89 & \NA & 42.45 & \NA \\
\abl{without question} & 30.86 & \textcolor{red!70!black}{-27.74} & 39.50 & \textcolor{red!70!black}{-16.95} & 53.08 & \textcolor{red!70!black}{-23.81} & 0.00 & \textcolor{red!70!black}{-42.45} \\
\abl{without image} & 21.45 & \textcolor{red!70!black}{-37.15} & 11.82 & \textcolor{red!70!black}{-44.63} & 50.70 & \textcolor{red!70!black}{-26.19} & 1.82 & \textcolor{red!70!black}{-40.63} \\
\addlinespace[3pt]
\midrule
Qwen3.5-397B-A17B (\textit{I}) & 58.25 & \NA & 45.26 & \NA & 77.59 & \NA & 51.91 & \NA \\
\abl{without question} & 30.55 & \textcolor{red!70!black}{-27.70} & 39.97 & \textcolor{red!70!black}{-5.29} & 51.68 & \textcolor{red!70!black}{-25.91} & 0.00 & \textcolor{red!70!black}{-51.91} \\
\abl{without image} & 26.25 & \textcolor{red!70!black}{-32.00} & 18.97 & \textcolor{red!70!black}{-26.29} & 50.00 & \textcolor{red!70!black}{-27.59} & 9.78 & \textcolor{red!70!black}{-42.13} \\
\addlinespace[3pt]
\midrule
Random & 16.79 & \NA & 0.38 & \NA & 50.00 & \NA & 0.00 & \NA \\
\bottomrule
\end{tabular}
}
\caption{Model accuracy by question type on the test split, under the full-input setting and two input ablations, with all values reported as percentages. Unindented rows report results with the complete input. Indented rows report results after removing either the image (\textit{without image}) or the question (\textit{without question}). For model variants, (\textit{T}) denotes \emph{Thinking} mode and (\textit{I}) denotes \emph{Instruct} mode.}
\label{tab:test-ablations}
\end{table}

\paragraph{Input ablations.}
Table~\ref{tab:test-ablations} reports performance under the full-input setting and after removing either the image or the question for top performing models. Removing the image reduces overall macro accuracy by 32.00--47.68 pp across models. Yes/no accuracy falls to 48.32--54.20\%, open-ended accuracy to 1.00--12.11\%, and multiple-choice accuracy to 4.35--29.08\%. The substantial performance drop after removing the image indicates that models rely heavily on visual information to answer questions in \ourDataset{}.

Removing the question, while retaining the image and mcq options, reduces open-ended accuracy to 0\% in reported models and leaves yes/no accuracy close to chance. Multiple-choice accuracy remains relatively high at 39.50--55.99\%, decreasing by only 5.29--16.95 points compared with the full-input setting. This shows that the image and answer options are often sufficient to identify the expected answer without access to the question. The result may reflect image--answer compatibility and differences in distractor plausibility.

\begin{figure}[t]
    \centering
    \includegraphics[width=\linewidth]{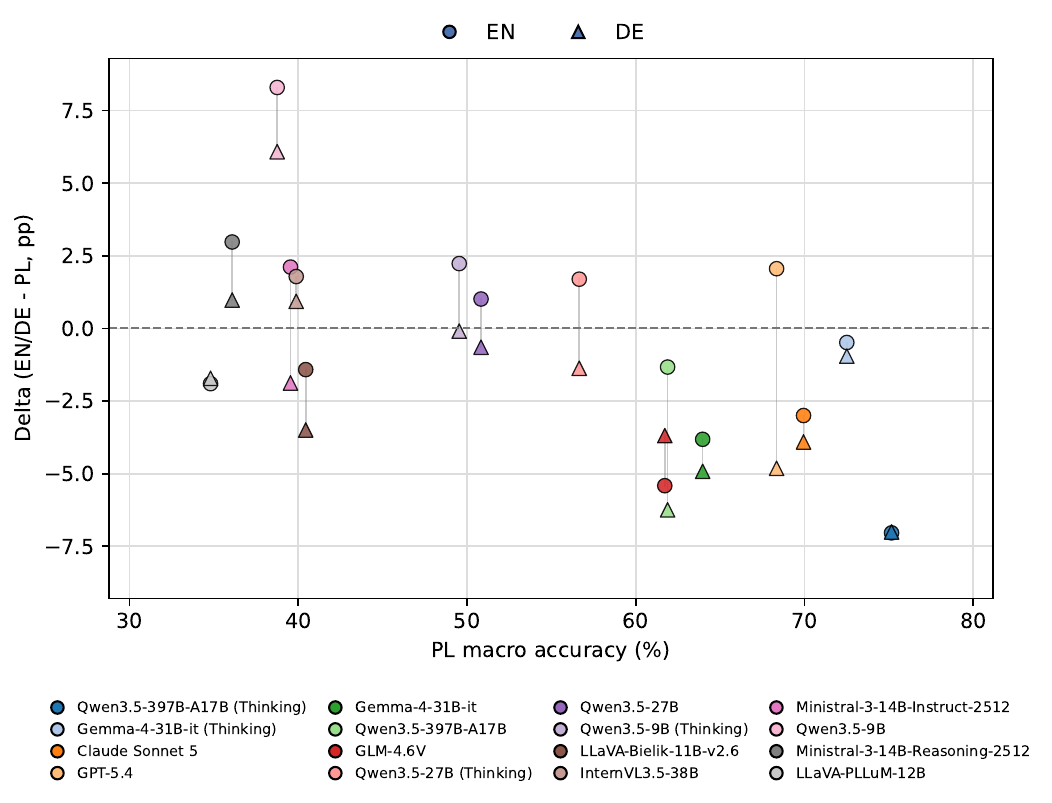}
    \caption{
    Translation effect by model. Macro accuracy in Polish (PL) is compared with the accuracy change after switching to English (EN) and German (DE). 
    }
    \label{fig:translation-effect-overall}
\end{figure}

\paragraph{Impact of prompt language.}
Figure~\ref{fig:translation-effect-overall} (detailed results in Table~\ref{tab:translation-overall-results} of Appendix~\ref{app:translation-results}) shows a relationship between performance on the original Polish questions and the effect of translation. Models with higher Polish accuracy generally perform worse when the questions are translated into English or German. Qwen3.5-397B-A17B \textit{Thinking}, the strongest model on the Polish subset, loses approximately 7 pp in both languages, while Claude Sonnet~5, Gemma-4-31B-it, and GLM-4.6V also show lower translated performance. In contrast, several lower-performing models improve with English prompts. The largest gain is observed for Qwen3.5-9B, which improves by 8.29 pp in English and 6.08 pp in German. This trend suggests that weaker general-purpose models may benefit from input in a language more strongly represented in their training data, whereas stronger models make better use of the original Polish formulation. GPT-5.4 is an exception, improving in English despite its high Polish performance, while the Polish-oriented LLaVA-Bielik and LLaVA-PLLuM models perform best in Polish.

\section{Related Work}
Early vision-language benchmarks primarily focused on image captioning, visual relations, object recognition and simple reasoning. Representative datasets such as MS COCO \cite{mscoco}, Visual Question Answering (VQA) v2 \cite{vqav2}, and CLEVR~\cite{Johnson_2017_CVPR} introduced tasks involving counting, spatial relations, and compositional reasoning (e.g., “How many objects are in the image?”). Subsequent benchmarks, including OK-VQA \cite{ok-vqa} and KVQA \cite{kvqa}, extended the evaluation towards external and common sense knowledge, while recent culturally grounded VQA benchmarks investigate whether models can interpret culturally specific symbols, practices, social norms, and geographically localized knowledge

More recently, a growing body of work has focused on evaluating vision-language models in culturally diverse settings. Large-scale multicultural benchmarks, such as CVQA \cite{cvqa}, cover dozens of countries and languages, enabling cross-cultural comparison. Similarly, CulturalVQA \cite{culturalvqa} and BlendVis \cite{blendvis} evaluate models across multiple geographic regions and categories, highlighting performance variability across cultures. The WorldCuisines dataset \cite{worldcuisines} further broadens cultural evaluation through food-related visual recognition tasks spanning multiple national cuisines, although Polish culture is represented only marginally within its coverage. However, these datasets prioritize breadth over depth, limiting their ability to capture fine-grained cultural understanding within a single context. In addition, some benchmarks rely on English-only questions, template-based generation, or synthetically generated images, which may introduce biases and conceptual inaccuracies undermining the value of the evaluation.

Another line of work focuses on region-specific benchmarks, including settings such as India \cite{drishtikon}, China and Taiwan \cite{cvlue, taiwanvqa}, and the Arab world \cite{jeem}. While more localized, these benchmarks often reflect substantial internal diversity, including multilingual and multidialectal variation, making controlled evaluation more challenging. Additionally, CVLUE, a Chinese vision-language understanding benchmark, focuses on image-level perception while addressing the Western-centric bias present in existing datasets, particularly in concept hierarchies derived from resources such as WordNet \cite{cvlue}. We include a comparison of \ourDataset{} with selected visual and culturally grounded benchmarks in Appendix~\ref{app:other_benchmarks}.

Despite the growing body of region-specific and multicultural benchmarks, Polish-language resources remain limited and fragmented.
The PLCC benchmark \cite{plcc} represents the first structured effort to evaluate culturally grounded knowledge in Polish, focusing on 600 text-only questions.
Similarly, LLMzSzŁ \cite{jassem2025llmzsz} introduces a large-scale evaluation framework for Polish language models based on a collection of national exams, covering nearly 19k closed-ended questions across multiple domains. While comprehensive, it remains restricted to the text modality and does not address multimodal understanding.
On the vision-language side, reVISION \cite{ciesiolka2025revision} provides a large-scale Polish benchmark for evaluating VLMs using questions derived from Polish national exams. 
However, its focus is primarily exam-driven and task-oriented, without explicitly modeling culturally grounded visual-linguistic competence or fine-grained cultural context.

To the best of our knowledge, there is no existing vision-language benchmark specifically designed for Polish cultural and linguistic evaluation beyond exam-based settings. In particular, current resources either focus on text-only evaluation (PLCC, LLMzSzŁ) or exam-centric multimodal reasoning (reVISION), leaving a gap in culturally grounded VLM benchmarks tailored to Polish context and everyday visual semantics.

\section{Conclusion and Future Work}

In this work, we introduced \ourDataset{}, a culturally grounded vision-language benchmark designed to evaluate multimodal models on Polish cultural and linguistic competence. The dataset combines manual, template-free annotation with a grounded evaluation paradigm, in which correct answers depend on the interaction between linguistic input and visual context.

Looking forward, our results highlight that open-ended questions constitute the most informative yet challenging evaluation setting, as they require models to generate precise, contextually grounded answers rather than select from predefined options. At the same time, they expose a fundamental challenge for evaluation. Exact-match scoring provides a transparent, reliable, and reproducible assessment framework, but it necessitates relatively constrained answer formats. We currently lack robust methodologies for reliably evaluating semantically equivalent yet linguistically diverse responses, particularly in culturally grounded settings. Developing such methods remains an important direction for future research.

\section*{Acknowledgements}

This work was supported by the Polish Ministry of Digital Affairs (subsidy no. 4/WII/DBI/2026).
The computational resources were provided by the Polish high-performance computing infrastructure PLGrid (HPC Center: ACK Cyfronet AGH) under computational grant no. PLG/2026/019138.

\section*{Limitations}

We acknowledge that constructing a culturally grounded benchmark of this nature involves inherent trade-offs between annotation quality, dataset size, and thematic coverage. Our decision to avoid template-based generation and synthetic data improves linguistic naturalness and cultural authenticity, but limits the scale of the dataset and may constrain the breadth of covered topics.

A central challenge lies in the evaluation of open-ended answers. While this sort of question offers the richest signal for assessing model capabilities, it is difficult to evaluate automatically using standard metrics. Approaches such as LLM-as-a-judge are particularly problematic in this setting, as they rely on models that may themselves exhibit cultural biases, creating a paradox when evaluating cultural competence. As a result, open-ended evaluation introduces both methodological complexity and potential bias.

Traditional evaluation methods based on string matching (e.g., exact match or token-level overlap) are insufficient for culturally grounded tasks, where multiple valid expressions may exist. To ensure reliable and reproducible evaluation, we therefore rely primarily on multiple-choice and binary formats. While these formats allow for deterministic evaluation, they also introduce their own limitations: models may succeed by eliminating implausible options rather than demonstrating genuine understanding. Although techniques such as circular evaluation help mitigate positional biases, they do not fully address this issue.

An additional challenge arises from the relationship between culturally prototypical expressions, annotator perspective, and conceptual variability. Annotators, as members of the target culture, naturally operate from an insider (emic) perspective and tend to favor culturally salient and widely shared labels. For example, certain architectural forms characteristic of the socialist era in Poland are commonly referred to as \textit{wielka płyta}, which functions as a culturally dominant prototype. However, the same phenomenon admits multiple valid descriptions, including more technical or general formulations (e.g., prefabricated housing structures), which may be preferred by experts or non-local observers.

This creates a perspective asymmetry, where culturally canonical answers are privileged over alternative, semantically correct interpretations. Furthermore, annotators may implicitly treat certain phenomena as culturally specific or unique, even when they are in fact shared across multiple regions. This may lead to over-localization, where questions assume a single culturally grounded interpretation despite the existence of broader or cross-cultural variants. As a result, models may be penalized for producing correct but non-prototypical answers, especially when these reflect alternative cultural, technical, or global perspectives.

Together, these effects highlight a fundamental challenge in culturally grounded evaluation: reconciling the need for precise and verifiable answers with the inherently plural, context-dependent nature of cultural and conceptual knowledge.

In our approach, we mitigate the impact of these biases by either reformulating questions to enforce a strict and unambiguous response format or by converting such instances into closed-ended questions. While this strategy improves evaluation consistency and reduces ambiguity, it introduces an inherent trade-off between linguistic and conceptual flexibility on the one hand and evaluation reliability on the other.

\FloatBarrier
\bibliography{custom}

\appendix

\section{Dataset examples and statistics visualization}
\label{app:examples}

\begin{figure*}[ht]
    \centering
    \includegraphics[width=\textwidth]{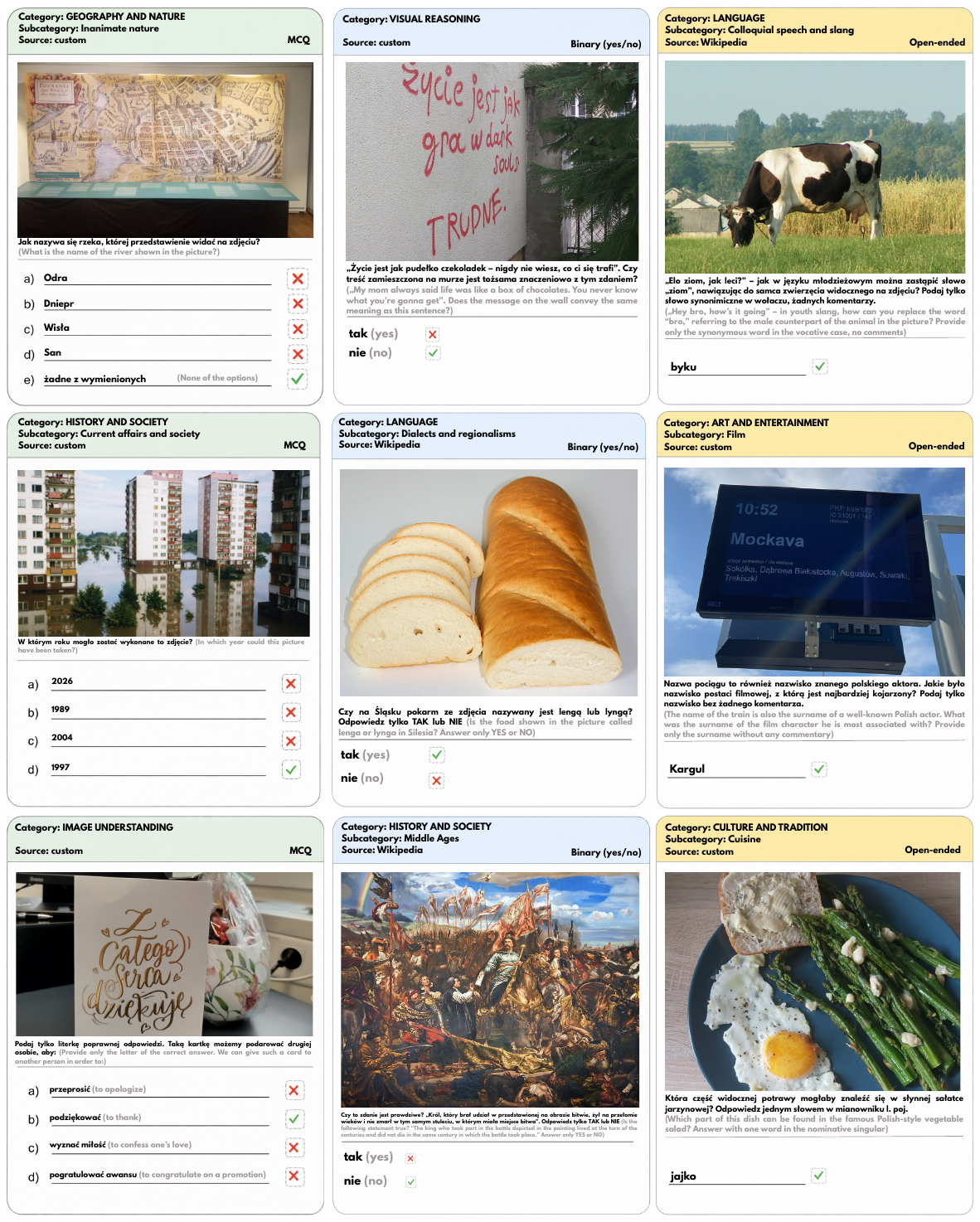}
    \caption{
Examples from the dataset illustrating multiple-choice, binary (yes/no), and open-ended questions across all seven main categories. Image sources include Wikimedia Commons and annotators' personal collections contributed to the project. English translations are omitted for Polish-specific named entities and for open-ended answers requiring Polish vocabulary.
}
    \label{fig:examples}
\end{figure*}

\begin{figure*}[t]
    \centering
    \includegraphics[width=0.95\textwidth]{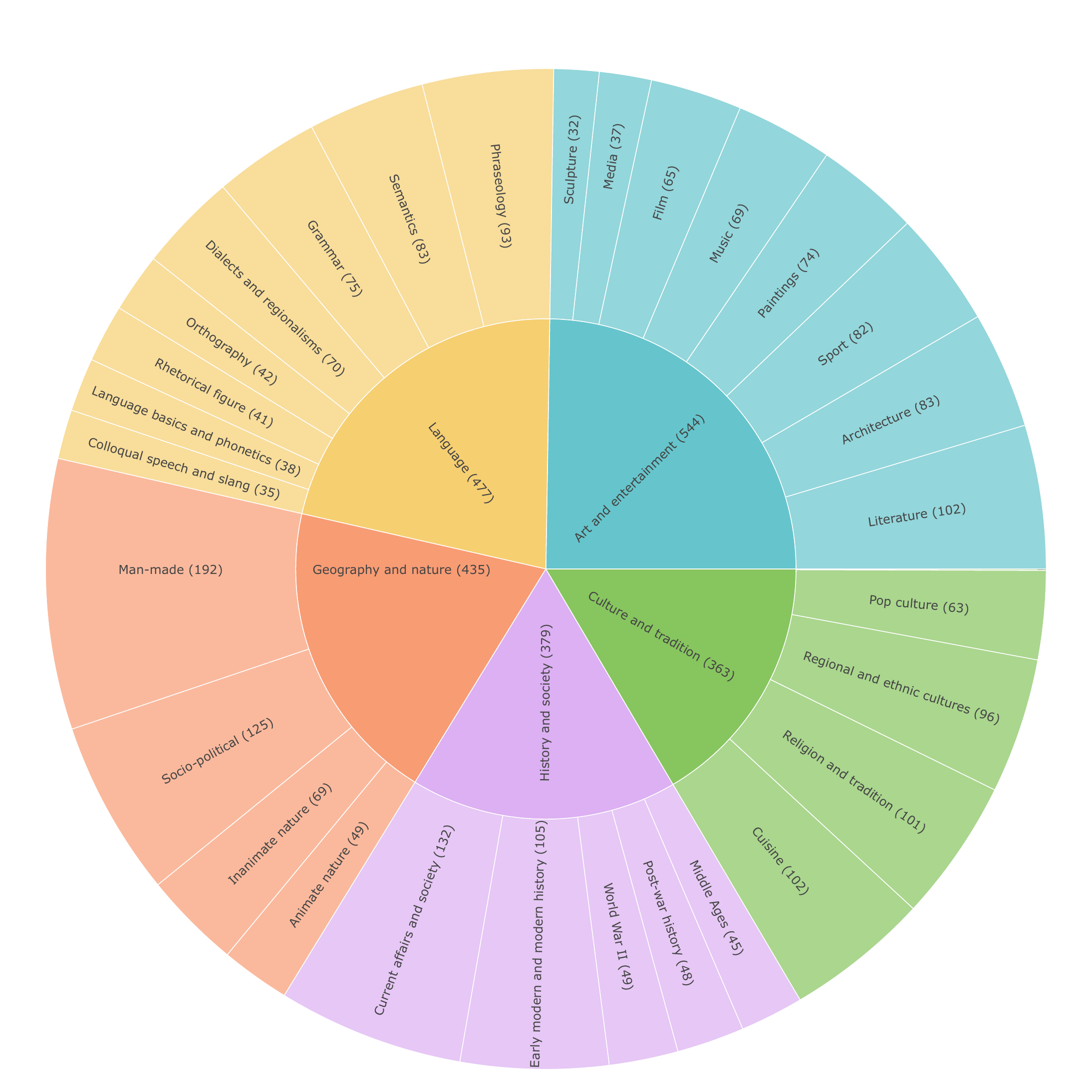}
    \caption{
    Distribution of 2{,}198 culturally grounded questions across categories and subcategories in the dataset.
    Numbers in parentheses indicate the number of samples.
    The figure excludes the additional categories \textit{Image Understanding} and \textit{Visual Reasoning} which are not divided into subcategories.
}
    \label{fig:sunburst}
\end{figure*}

Figure~\ref{fig:examples} presents representative examples from the dataset, illustrating the diversity of question types and linguistic formulations, while Figure~\ref{fig:sunburst} visualizes this distribution in a hierarchical form, offering an intuitive overview of the balance between high-level categories and their internal structure.

\clearpage

\section{Linguistic Analysis of Questions}
\label{stylometric}

To quantify the linguistic properties of the dataset, we focus exclusively on questions, as answers are typically very short (one to two tokens), limiting their usefulness for stylometric analysis. For MCQ tasks, options provided are excluded and not considered part of the question.

\begin{figure*}[h]
    \centering
    \includegraphics[width=0.8\textwidth]{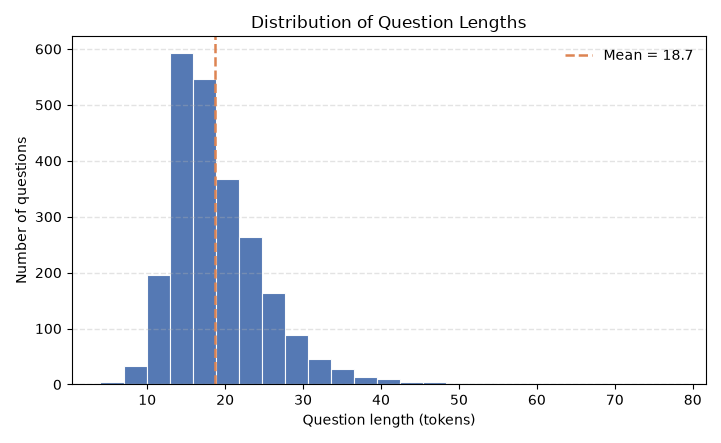}
    \caption{
Distribution of question lengths (in tokens) across the dataset. The dashed vertical line indicates the mean question length (18.7 tokens).
}
    \label{fig:histogram}
\end{figure*}

The average question length is 18.7 tokens (Figure~\ref{fig:histogram}). The distribution is right-skewed, with most questions containing between approximately 12 and 25 tokens and a long tail extending toward longer questions. This indicates that while the majority of questions are of moderate length, a smaller number of substantially longer questions are also present in the dataset.

\subsection{Stylometric features}
To further investigate the linguistic nature of the dataset, we employed StyloMetrix \cite{okulska2023}, a library designed for Polish, which represents texts as vectors of interpretable linguistic features normalized to the range $[0,1]$. We focused in particular on metrics capturing lexical diversity and syntactic structure, including indicators associated with the presence of subordinate clauses, which may signal multi-hop reasoning.  Selected metrics are reported in Table~\ref{tab:stylometry}.

Interestingly, we observe a minimal difference between the surface-form type–token ratio (L\_TTR\_IA) and its lemmatized counterpart (L\_TTR\_LA) (0.850 vs.\ 0.841), despite the fact that many questions follow similar, though not identical, task formulations designed to ensure unambiguous evaluation of model behavior.This suggests that lexical diversity is not primarily driven by morphological variation, which is characteristic of richly inflected Slavic languages, but instead reflects genuinely diverse vocabulary usage. 

Furthermore, the combination of a named entity ratio (0.031) and a high type–token ratio indicates that entity mentions are not dominated by a small set of frequently repeated references, but instead span a broad range of distinct entities. This contributes to the dataset’s topical and regional diversity. This observation is reinforced by nearly identical content word incidence and content word types (0.5996 vs.\ 0.5967), suggesting that many content words occur only once and thus contribute directly to lexical diversity.

The proportion of pronouns in the dataset is comparable to that of adjectives (0.099 vs. 0.096), indicating that reference to entities is frequently realized through pronominal forms rather than descriptive modification. Relative and interrogative pronouns further contribute to the prevalence of subordinate structures, while the presence of negative pronouns reflects the inclusion of adversarial question formulations.

Despite the expectation that a question dataset should consist predominantly of interrogative sentences, our dataset contains a relatively high proportion of tokens associated with declarative constructions (0.626 vs. 0.349). This can be partly attributed to strict task formulations, which are often expressed in short declarative phrases. 

However, these formulations alone do not account for the observed distribution. The predominance of declarative tokens is further driven by the frequent use of contextual or narrative framing preceding the actual question. While the interrogative component itself may be relatively short, it is often embedded within a longer descriptive context, resulting in a significant proportion of declarative structures. A smaller, yet non-negligible proportion of tokens corresponds to negative constructions (0.035), indicating the presence of adversarial or contrastive question formulations.

Finally, high values for words within modifiers (0.296) and words in nominal phrases (0.613) indicate a strongly periphrastic style of question formulation. This style likely supports the construction of indirect, image-grounded references and enables the formulation of questions that require interpretation beyond explicitly depicted visual elements.

\subsection{Question-level syntactic complexity}
To further illustrate the distribution of complexity across the questions and having thoroughly examined stylometric features extracted with StyloMetrix, we aimed to quantify syntactic complexity, which may indicate the need for multi-hop reasoning and contribute to the overall linguistic difficulty of the questions. To this end, we introduce three custom metrics based on rule-based detection of subordinate structures: (i) Relative Clause Proportion (RCP), (ii) Subordinate Conjunction Proportion (SCP), and (iii) Subordination Proportion (SP), which captures the presence of either structure.

Each metric is computed at the question level as the proportion of questions containing at least one instance of the corresponding structure. For all metrics, sentence-initial tokens (including capitalized pronouns and conjunctions) are excluded, as they typically correspond to interrogative openings rather than embedded subordinate constructions.

To further characterize the distribution of structural complexity across the questions, and building on the stylometric analysis performed with StyloMetrix, we aimed to quantify syntactic complexity, which may indicate the need for multi-hop reasoning and contribute to the overall linguistic difficulty of the dataset. 

To this end, we introduce three custom metrics based on rule-based detection of subordinate structures: (i) Relative Clause Proportion (RCP), (ii) Subordinate Conjunction Proportion (SCP), and (iii) Subordination Proportion (SP), which captures the presence of either structure.

Each metric is computed at the question level as the proportion of questions containing at least one instance of the corresponding structure. For all metrics, sentence-initial tokens (including capitalized pronouns and conjunctions) are excluded, as they typically correspond to interrogative openings rather than embedded subordinate constructions.

The resulting values are as follows: RCP = 0.334, SCP = 0.081, and SP = 0.384. This also reveals an overlap between the two indicators, corresponding to questions that contain both a relative clause and a subordinate conjunction (3.09\% of the dataset), indicating the presence of more complex, multi-layered clause structures. Importantly, these metrics are computed as sentence-level incidence proportions rather than token-level ratios. This design choice is motivated by the presence of task formulations, which could otherwise inflate token-based measurements and obscure the true distribution of structurally complex questions. 

Overall, the dataset exhibits high lexical diversity, rich semantic content, and structurally complex, multi-clause question formulations, indicating that it is not dominated by simple templates but supports more demanding, context-driven reasoning.

\begin{table}[h]
\centering
\small
\setlength{\tabcolsep}{4pt}
\begin{tabular}{lp{3.5cm}r}
\toprule
\textbf{Metric} & \textbf{Description} & \textbf{Score} \\
\midrule

\multicolumn{3}{l}{\textbf{Grammatical Forms}} \\
G\_N & Nouns & 0.278 \\
G\_V & Verbs & 0.128 \\
G\_ADJ & Adjectives & 0.096 \\
G\_ADV & Adverbs & 0.027 \\
G\_PRO & Pronouns & 0.099 \\
G\_PRO\_NEG & Negative pronouns & 0.003 \\
G\_PRO\_REL & Relative pronouns & 0.009 \\
G\_PRO\_INT & Interrogative pronouns & 0.034 \\
G\_CONJ & Conjunctions & 0.029 \\
G\_CCONJ & Coordinating conjunctions & 0.026 \\
G\_SCONJ & Subordinating conjunctions & 0.003 \\

\midrule
\multicolumn{3}{l}{\textbf{Punctuation}} \\
PUNCT\_TOTAL & Total punctuation & 0.133 \\

\midrule
\multicolumn{3}{l}{\textbf{Syntactic}} \\
SY\_MOD & Words within modifiers & 0.296 \\
SY\_NPHR & Words in nominal phrases & 0.613 \\
SY\_S\_DE & Words in declarative sentences & 0.349 \\
SY\_S\_IN & Words in interrogative sentences & 0.626 \\
SY\_S\_NEG & Words in negative sentences & 0.035 \\
SY\_QUOT & Words in quotation marks & 0.001 \\

\midrule
\multicolumn{3}{l}{\textbf{Lexical}} \\
L\_TTR\_IA & Type-token ratio for non-lemmatized tokens & 0.850 \\
L\_TTR\_LA & Type-token ratio for lemmatized tokens & 0.841 \\
L\_CONT\_A & Incidence of content words & 0.600 \\
L\_CONT\_T & Content word types & 0.600 \\
L\_FUNC\_A & Incidence of function words & 0.242 \\
L\_FUNC\_T & Function words types & 0.233 \\
L\_NAME\_ENT & Named entities & 0.031 \\
L\_NAME & Proper names & 0.015 \\
L\_PERSN & Person names & 0.005 \\
L\_PLACEN\_GEOG & Place and geographical names & 0.009 \\

\bottomrule
\end{tabular}
\caption{Stylometric feature distribution of the dataset across grammatical, syntactic, and lexical categories.}
\label{tab:stylometry}
\end{table}

\section{Comparison to Other Datasets}
\label{app:other_benchmarks}

Table~\ref{tab:dataset_comparison} compares \ourDataset{} with existing culturally grounded VQA benchmarks across scale, geographic coverage, and key dataset properties. While some prior datasets achieve larger scale by covering multiple countries or regions, \ourDataset{} adopts a monocultural design focused on a single, well-defined context. This allows for a more controlled and fine-grained evaluation of cultural and linguistic competence. In particular, we emphasize fully manual, creative question construction, the use of a local (non-English) language, and the inclusion of non-synthetic visual data, including images sourced from annotators’ private collections that are not publicly available online.

We additionally include CUS-QA in the comparison. While it is primarily knowledge-oriented, it is embedded in the Czech, Slovak, and Ukrainian context, which is relevant due to the geographic and cultural proximity of these regions. 

When reporting dataset statistics, we provide numbers to the best of our knowledge, focusing specifically on the visual question answering setting. In cases where datasets support multiple evaluation formats, only the visual components are considered to ensure comparability. The only exception is PLCC, which we include as a baseline for Polish cultural understanding; notably, it is a text-only benchmark.

We decided not to include the Polish Cultural Vision Benchmark (PCVB) v2 in this comparison.\footnote{\url{https://huggingface.co/spaces/speakleash/Polish_Cultural_Vision_Benchmark}} While it is presented as a benchmark for Polish cultural understanding in vision-language models, as of August 2026 no documentation regarding its construction, annotation methodology, or evaluation protocol is publicly available, which prevents a meaningful and reproducible comparison.

\begin{table*}[h]
\centering
\footnotesize
\setlength{\tabcolsep}{4pt}
\renewcommand{\arraystretch}{0.95}
\resizebox{\linewidth}{!}{
\begin{tabular}{l p{2.5cm} r r l c c c c c}
\toprule
\textbf{Dataset} & \textbf{Regions} & \textbf{\#Img} & \textbf{\#Q} & \textbf{Lang.} & \textbf{Manual} & \textbf{MCQ} & \textbf{Open} & \textbf{Own Img.} & \textbf{Synth Img.} \\
\midrule

CULTURALVQA & 
11 (global)
& 2{,}328 & 2{,}378 
& EN 
& \textcolor{OliveGreen}{\cmark} &  \textcolor{red}{\xmark} & \textcolor{OliveGreen}{\cmark} & \textcolor{red}{\xmark} & \textcolor{red}{\xmark} \\

\midrule

BLEND-VIS & 
16 (global)
& 4{,}916 & 21{,}782 
& EN 
& \textcolor{red}{\xmark} & \textcolor{OliveGreen}{\cmark} & \textcolor{red}{\xmark} & \textcolor{red}{\xmark} & \textcolor{OliveGreen}{\cmark} \\

\midrule

DRISHTIKON & 
1 (IN)
& 2{,}126 & 64{,}288 
& EN + dialects 
& \textcolor{red}{\xmark} & \textcolor{OliveGreen}{\cmark} & \textcolor{red}{\xmark} & \textcolor{red}{\xmark} & \textcolor{red}{\xmark} \\

\midrule

JEEM & 
4 (MENA)
& 2{,}178 & {10{,}890}
& AR 
& \textcolor{OliveGreen}{\cmark} & \textcolor{red}{\xmark} & \textcolor{OliveGreen}{\cmark} & \textcolor{OliveGreen}{\cmark} & \textcolor{red}{\xmark} \\

\midrule

CVLUE & 
1 (CN)
& \textcolor{gray}{30{,}009 (?)} & \textcolor{gray}{72{,}306 (?)} 
& ZH 
& \textcolor{OliveGreen}{\cmark} & \textcolor{red}{\xmark} & \textcolor{OliveGreen}{\cmark} & \textcolor{red}{\xmark} & \textcolor{red}{\xmark} \\

\midrule

TaiwanVQA & 
1 (TW)
& 2{,}736 & 5{,}472 
& ZH 
& \textcolor{OliveGreen}{\cmark} & \textcolor{OliveGreen}{\cmark} & \textcolor{OliveGreen}{\cmark} & \textcolor{OliveGreen}{\cmark} & \textcolor{red}{\xmark} \\

\midrule

CUS-QA & 
3 (CZ, SK, UA) 
& \textcolor{gray}{(?)} & 1,097 
& CZ, SK, UA 
& \textcolor{OliveGreen}{\cmark} & \textcolor{red}{\xmark}  & \textcolor{OliveGreen}{\cmark}  & \textcolor{red}{\xmark}  & \textcolor{red}{\xmark}  \\

\midrule

Afri-MCQA & 
12 (African countries) 
& \textcolor{gray}{3,000 (?)} & 7,642 
& EN + 15 African languages 
& \textcolor{OliveGreen}{\cmark} & \textcolor{OliveGreen}{\cmark}  & \textcolor{OliveGreen}{\cmark}  & \textcolor{OliveGreen}{\cmark}  & \textcolor{red}{\xmark}  \\

\midrule

PLCC & 
1 (PL)
& --- & 600 
& PL 
& \textcolor{OliveGreen}{\cmark} & \textcolor{OliveGreen}{\cmark} & \textcolor{OliveGreen}{\cmark} & --- & --- \\

\midrule

\textbf{\ourDataset{}} &
\textbf{1 (PL)} &
\textbf{1{,}117} &
\textbf{2{,}366} &
\textbf{PL} &
\textbf{\textcolor{OliveGreen}{\cmark}} &
\textbf{\textcolor{OliveGreen}{\cmark}} &
\textbf{\textcolor{OliveGreen}{\cmark}} &
\textbf{\textcolor{OliveGreen}{\cmark}} &
\textbf{\textcolor{red}{\xmark}} \\

\bottomrule
\end{tabular}
}
\caption{Comparison of culturally grounded VQA datasets and culture-aware benchmarks. While most listed resources are visual question answering (VQA) datasets, we additionally include PLCC as a text-only baseline for Polish culture-aware evaluation. Regions are reported as the number of covered countries or regions with indicative geographic scope. Checkmarks indicate key dataset properties, including annotation type, use of annotator-provided (non-web) images, and the presence of synthetic images.}

\label{tab:dataset_comparison}
\end{table*}

\section{Annotation Guidelines}
\label{sec:guidelines}
\subsection{Manual data collection and annotation}
The aim of the dataset annotation is to manually craft questions regarding visual content defined by the following categories:

\begin{itemize}

\item \textbf{Art \& Entertainment} – artistic, cultural, and media-related content.
    \begin{itemize}
        \item \textbf{Architecture} – famous buildings, monuments, and structural design.
        \item \textbf{Film} – movies, scenes, actors, and film-related references.
        \item \textbf{Literature} – books, authors, and literary history.
        \item \textbf{Media} – press, television, and digital media.
        \item \textbf{Music} – musical works, performers, and related cultural references.
        \item \textbf{Paintings} – paintings, artists, and related artistic phenomena.
        \item \textbf{Sculpture} – sculptures, sculptors, and related artistic works.
        \item \textbf{Sport} – sports activities, events, and figures.
    \end{itemize}

\item \textbf{Geography \& Nature} – physical environment and spatial context.
    \begin{itemize}
        \item \textbf{Animate nature} – animals and living organisms.
        \item \textbf{Inanimate nature} – landscapes and natural formations.
        \item \textbf{Man-made} – human-made geographical, urban, industrial, and infrastructural elements, including landmarks.
        \item \textbf{Socio-political} – regions, borders, and administrative entities.
    \end{itemize}

\item \textbf{Culture \& Tradition} – customs, practices, and shared cultural symbols.
    \begin{itemize}
        \item \textbf{Cuisine} – food, beverages, and culinary traditions.
        \item \textbf{Pop culture} – contemporary cultural trends and references.
        \item \textbf{Regional and ethnic cultures} – local traditions and cultural variation.
        \item \textbf{Religion and tradition} – rituals, beliefs, and heritage practices.
    \end{itemize}

\item \textbf{History \& Society} – historical events and social context.
    \begin{itemize}
        \item \textbf{Middle Ages} – medieval historical period.
        \item \textbf{Early modern and modern history} – post-medieval developments to 20th century.
        \item \textbf{World War II} – events and figures related to WWII.
        \item \textbf{Post-war history} – developments after 1945.
        \item \textbf{Current affairs and society} – contemporary social and political issues.
    \end{itemize}

\item \textbf{Language} – linguistic structure, meaning, and variation.
    \begin{itemize}
        \item \textbf{Colloquial speech and slang} – informal language use.
        \item \textbf{Dialects and regionalisms} – region-specific language variation.
        \item \textbf{Grammar} – syntactic structure and correctness.
        \item \textbf{Language basics and phonetics} – pronunciation, sounds, and basic elements of language.
        \item \textbf{Orthography} – spelling and writing conventions.
        \item \textbf{Phraseology} – idioms and fixed expressions.
        \item \textbf{Rhetorical figures} – stylistic and figurative language.
        \item \textbf{Semantics} – word meaning and interpretation.
    \end{itemize}

\item \textbf{Image Understanding} – direct recognition and interpretation of visual content.

\item \textbf{Visual Reasoning} – reasoning about relationships and context within an image.

\end{itemize}

\vspace{0.5em}

\noindent
The annotator’s task consists of the following steps:

\begin{enumerate}
    \item \textbf{Image selection:} Choose an image from one of the following sources:
    \begin{itemize}
        \item Wikimedia Commons
        \item Publicly available sources (after careful verification of the license terms)
        \item Personal image collections
    \end{itemize}

    \item \textbf{Image upload and metadata annotation:} Upload the selected image to the annotation interface and provide the required metadata, including:
    \begin{itemize}
        \item A source hyperlink, or
        \item A declaration confirming ownership of the image and agreement to share it for research purposes under the CC BY-SA license
    \end{itemize}

    \item \textbf{VQA pair creation:} Create from one up to ten question–answer (VQA) pairs based on the image content.

    \item \textbf{Annotation labels:}
    \begin{itemize}
        \item \textbf{Task type:} multiple-choice, binary (yes/no), open-ended questions
        \item \textbf{Category:} category and subcategory
    \end{itemize}
\end{enumerate}

Follow the rules and guidelines for annotation:
\begin{itemize}
    \item \textbf{Formulation of questions:} Construct natural-sounding questions that reflect authentic language use. Ensure linguistic correctness and fluency. Avoid repetitive or overly uniform question structures; instead, aim for diversity in formulation and richness of vocabulary.

    \item \textbf{Answer unambiguity:} Ensure that all answers are unambiguous and allow for an objective and verifiable evaluation. In the case of multiple-choice questions (MCQ), the distinction between correct and incorrect options must be clear and indisputable. Avoid answer choices that could be interpreted as correct depending on regional variation, interpretation, or context (e.g., cases where answer A may be valid in one region while answer B is accepted elsewhere). For open-ended questions, aim to formulate them in such a way that only one answer is clearly plausible. Where appropriate, include subtle guidance in the question (e.g., expected format or level of specificity) to reduce variability in answers. If in doubt regarding the ambiguity of a VQA pair, consult with the super-annotator or discuss it with the team to ensure consistency and agreement across annotators.

    \item \textbf{Coverage and difficulty:} Ensure that the dataset covers a broad range of thematic categories and includes varying levels of difficulty. Questions may rely on cultural and region-specific knowledge that is broadly accessible through public education, media, and cultural institutions. However, they should not require specialized academic knowledge or expert-level training.
    
    \item \textbf{Dependence on visual content:} All questions must require image understanding. The answer should not be obtainable from textual knowledge alone without reference to the image.

    \item \textbf{Image-referential language:} Use image-grounded expressions (e.g., “the man in the picture”) rather than overly specific or leading descriptions (e.g., “the famous musician in the picture”). Avoid including clues that reveal the answer directly.

    \item \textbf{Reasoning and multi-hop questions:} Whenever possible, formulate questions that require multi-step reasoning, combining multiple pieces of information from the image or integrating visual cues with culturally grounded or general knowledge. Such questions should go beyond direct recognition and involve interpretation, inference, or linking distinct visual elements.

    \textit{Example:} “Which sport is this man's wife known for?”

    \textit{Interpretation:} This question constitutes a multi-hop reasoning task when the sport is not directly observable in the image and the model must: (i) recognize the man (e.g., a public figure), (ii) infer the identity of his spouse, and (iii) recall the sport discipline in which she is active.

    \item \textbf{Use of associations:} Treat the image as a visual anchor that can connect to broader cultural or knowledge contexts. 
    Questions may refer to entities or concepts associated with the image content rather than only to what is directly visible. However, the image must provide the essential context needed to identify the relevant entity, object, place, or event and answer the question.

    \item \textbf{Adversarial questions:} Include a subset of carefully designed adversarial questions to evaluate models' robustness. These questions should be intentionally challenging, e.g., involving misleading visual cues, while still remaining answerable based on the image.

    \item \textbf{Design of incorrect answer options:} In multiple-choice questions (MCQ), ensure sufficient variation and plausibility among incorrect answers. For lower-difficulty questions, distractors (incorrect options) may be more similar to the correct answer, requiring careful distinction. For higher-difficulty questions, distractors may be less subtle; however, they should remain relatively plausible. Avoid overly absurd or clearly incorrect options, as these allow the correct answer to be identified through simple elimination without reference to the image.
    
\end{itemize}

\subsection{Augmented Data Annotation}
\subsubsection{Image Filtering and Annotation}
At the stage of augmented data annotation the task is to decide whether the pre-collected set of images can be annotated in consistence with the established guidelines and insights from the previous stage of manual annotation. The set of images contains different Wikimedia-based images retrieved from the Poland-related categories. 

The annotator’s task is to:
\begin{itemize}
    \item Annotate the VQA pair and assign the appropriate task type and category, following the procedures defined in the first stage.
    
    \item If the image is rejected, assign one of the following labels:
    \begin{itemize}
        \item Low quality
        \item Not suitable for VQA
        \item Potentially relevant but outside my domain of expertise
    \end{itemize}
\end{itemize}

\subsubsection{Question-Guided Image Matching}
At the final stage of the augmented data annotation process, annotators are tasked with identifying images that match previously created questions. Since multiple questions may be associated with a single source image, this step aims to diversify the dataset by retrieving visually similar images from Wikimedia Commons for further human inspection. The goal is to distribute questions across a broader set of non-identical images whenever possible.

The matching procedure is based on the previously constructed annotations (image–question pairs) and follows these guidelines:

\begin{itemize}
    \item Verify whether a given question can be answered using any of the automatically retrieved candidate images other than the original one. If so, mark the image as a valid match.
    \item Note that some candidate images may be near-duplicates (e.g., slightly different crops or framing). Such cases should not be treated as distinct images.
    \item For artworks such as paintings, partial views of the original canvas may be labeled as valid matches, provided that they contain sufficient information to answer the question.
    \item If none of the candidate images are applicable, submit the task and proceed to the next sample.
\end{itemize}

\subsection{Cross-validation}

At this stage, each annotated instance is reviewed by another annotator through careful manual inspection, following a structured set of validation criteria.

\begin{itemize}
    \item \textbf{Linguistic and logical correctness:} Verify that the question is grammatically correct, natural in phrasing, and logically well-formed, i.e., it is internally consistent, does not contain contradictory or ill-defined references, and clearly specifies the entity or concept being asked about.
    \item \textbf{Answer correctness and unambiguity:} Ensure that the answer is factually accurate and unequivocal, with no plausible alternative interpretations. Ensure that the answer can be inferred from the image and any relevant contextual or textual clues provided in the question. Consider potential biases arising from regional or cultural assumptions, and verify whether the interpretation is broadly understandable; if it relies on specific local knowledge, either reformulate the question or make the required context explicit (e.g., when referring to a regional term, indicate that it may be non-standard or region-specific).
    \item \textbf{Visual grounding:} Ensure that the answer can be derived solely from the visual content and that the question cannot be answered without access to the image.
    \item \textbf{Compliance with VQA criteria:} Confirm that the question adheres to the defined task requirements, including its Polish cultural relevance and appropriate category assignment.
    \item \textbf{Metadata verification:} Ensure that all required metadata is complete and that licensing terms have been correctly applied. In particular, verify that no copyrighted material is included in violation of the dataset’s licensing constraints.
\end{itemize}

If any issues or inaccuracies are identified, apply appropriate revisions:
    \begin{itemize}
        \item Reformulate the question and the answer for logical clarity or linguistic correctness
        \item Adjust the question type if necessary
        \item Replace the question if it does not meet the guidelines
        \item Delete the image or question if it is found to be non-compliant with VQA criteria
    \end{itemize}

\textbf{Escalation:} In cases of uncertainty, consult the super-annotator to ensure consistency with the overall annotation standards. Any full replacement or deletion must be explicitly approved by the super-annotator.

\textbf{Super-annotator interaction:} Be aware that the super-annotator may review selected samples throughout the annotation process. Annotators are expected to incorporate their feedback, including suggested corrections, reformulations, and improvements. In particular, annotators may be asked to revise questions that exhibit recurring issues, lack clarity, or reflect systematic inconsistencies, in order to maintain overall dataset quality and consistency.

\subsection{Iterative Quality Control} 

As part of the iterative quality control process, annotators will receive dedicated review sheets containing the results of validation analyses conducted on a selected subset of 11 LLMs. These sheets serve as supplementary material and identify VQA instances that may require revision. Any necessary changes should be implemented directly in the annotation tool. 
\begin{itemize} 
\item \textbf{Multiple-choice question difficulty assessment:} Review VQA instances flagged on the basis of image-and-options-only evaluations, in which LLMs were provided with the image and answer options but not the corresponding question. Questions for which the models achieved a high success rate (approximately above 60\%) should be considered potentially too easy. When revising such questions, possible actions include: 
\begin{itemize} 
\item replacing distractors with more plausible/challenging alternatives, 
\item increasing the number of answer options, 
\item adding alternatives such as "none of the options" when appropriate, \item converting the question into an open-ended format when a single clear and unambiguous answer can reasonably be expected. 
\end{itemize} 
\item \textbf{Visual grounding assessment:} Review questions flagged on the basis of text-only evaluations, in which LLMs were given the question without access to the image. Questions that can be answered correctly without visual information should be considered insufficiently grounded in the image content. Such questions should be reformulated, replaced, or removed. 
\item \textbf{Open-ended answer validation:} Analyze review sheets containing open-ended questions, LLM-generated responses, and the corresponding automated evaluation results. Each sample should be inspected manually to verify that correct answers are consistently scored as correct and that incorrect answers are not accepted by the evaluation protocol. When reviewing a sample, take into account the following: 
\begin{itemize} 
\item identify plausible alternative correct answers that are not currently accepted and add them to the list of accepted responses when appropriate, 
\item verify compliance with grammatical and orthographic requirements, 
\item ensure that answers containing orthographic errors are not accepted, even if they refer to the correct entity or event, 
\item reformulate the question if the current evaluation protocol does not allow for unambiguous assessment of responses, 
\item check for ambiguities or inconsistencies in the answer-matching rules that go beyond the adjustments annotators can make and may lead to incorrect scoring; such issues should be reported to the evaluation team so that the matching rules can be revised accordingly. 
\end{itemize}
\end{itemize}

\subsection{Annotators' demographics}
\label{demographics}

The annotation team consisted of 16 participants representing four age groups. The distribution was balanced across age categories, with four annotators (25\%) in each group: 20--25, 25--30, 30--35, and 35--40 years. All annotators have resided in Poland. 

Annotators were assigned one of three roles. Two \textit{Primary} annotators were responsible for the core annotation process, one \textit{Super-annotator} oversaw annotation quality and guideline compliance, and the remaining thirteen \textit{Auxiliary} annotators contributed by supplying their own images as candidate visual question answering (VQA) instances, thereby increasing the regional diversity of the dataset.

The annotators originated from 11 of Poland's 16 voivodships, providing broad geographic coverage. In addition to their region of origin, annotators reported a \textit{current/familiarized region}, defined as a voivodship in which they had lived, studied, or worked for a substantial period of time. This distinction allowed us to account not only for birthplace but also for regional familiarity acquired through migration and long-term residence. The demographic characteristics of the annotation team are presented in Table~\ref{tab:annotator_demographics}.

\begin{table*}[tbh]
\centering
\small
\begin{tabular}{lllll}
\toprule
Annotator ID & Role & Age Group & Voivodship of Origin & Current/Familiarized Region\\
\midrule
P1  & Primary         & 20--25 & opolskie             & mazowieckie \\
P2  & Primary         & 35--40 & małopolskie          & opolskie \\
S1  & Super-annotator & 35--40 & dolnośląskie         & wielkopolskie \\
A1  & Auxiliary       & 20--25 & dolnośląskie         & dolnośląskie \\
A2  & Auxiliary       & 25--30 & podlaskie            & mazowieckie \\
A3  & Auxiliary       & 20--25 & wielkopolskie        & wielkopolskie \\
A4  & Auxiliary       & 25--30 & mazowieckie          & mazowieckie \\
A5  & Auxiliary       & 35--40 & pomorskie            & pomorskie \\
A6  & Auxiliary       & 30--35 & świętokrzyskie       & mazowieckie \\
A7  & Auxiliary       & 30--35 & lubelskie            & mazowieckie \\
A8  & Auxiliary       & 30--35 & śląskie              & mazowieckie \\
A9  & Auxiliary       & 30--35 & mazowieckie          & mazowieckie \\
A10 & Auxiliary       & 25--30 & dolnośląskie         & mazowieckie \\
A11 & Auxiliary       & 35--40 & podlaskie            & mazowieckie \\
A12 & Auxiliary       & 20--25 & warmińsko-mazurskie  & mazowieckie \\
A13 & Auxiliary       & 25--30 & dolnośląskie         & mazowieckie \\
\bottomrule
\end{tabular}
\caption{Demographic characteristics of the annotation team.}
\label{tab:annotator_demographics}
\end{table*}

\section{Implementation Details}
\label{app:computational}

Below we provide details on specific model versions, prompts, decoding parameters and computational costs.

\subsection{Model Versions}

Table~\ref{tab:model-versions} lists the exact model versions and identifiers used in our experiments. Open-weight models were served locally with vLLM, whereas proprietary models were accessed through OpenRouter. We report the corresponding repository or detail model version in API.

\begin{table*}[tbh]
\centering
\small
\setlength{\tabcolsep}{4pt}
\renewcommand{\arraystretch}{1.15}

\begin{tabularx}{\textwidth}{@{}
    >{\raggedright\arraybackslash}p{0.23\textwidth}
    >{\raggedright\arraybackslash}p{0.055\textwidth}
    >{\raggedright\arraybackslash}p{0.11\textwidth}
    >{\raggedright\arraybackslash}p{0.10\textwidth}
    >{\centering\arraybackslash}p{0.07\textwidth}
    >{\scriptsize\ttfamily\raggedright\arraybackslash}X
@{}}
\toprule
Model & Size & Type & Backend & Thinking &
{\normalfont\small ID} \\
\midrule

GPT-5.4
    & -- & proprietary & OpenRouter & 
    & \nolinkurl{openai/gpt-5.4-20260305} \\

Claude Sonnet 5
    & -- & proprietary & OpenRouter & 
    & \nolinkurl{anthropic/claude-sonnet-5-20260630} \\

Qwen3.5-9B
    & 9B & open weight & vLLM & \textbf{\textcolor{OliveGreen}{\cmark}}
    & \nolinkurl{Qwen/Qwen3.5-9B} \\

Qwen3.5-27B
    & 27B & open weight & vLLM & \textbf{\textcolor{OliveGreen}{\cmark}}
    & \nolinkurl{Qwen/Qwen3.5-27B} \\

Qwen3.5-397B-A17B
    & 397B & open weight & vLLM & \textbf{\textcolor{OliveGreen}{\cmark}}
    & \nolinkurl{Qwen/Qwen3.5-397B-A17B} \\

Gemma-4-31B-it
    & 31B & open weight & vLLM & \textbf{\textcolor{OliveGreen}{\cmark}}
    & \nolinkurl{google/gemma-4-31B-it} \\

Mistral-Medium-3.5-128B
    & 128B & open weight & vLLM & 
    & \nolinkurl{mistralai/Mistral-Medium-3.5-128B} \\

Ministral-3-14B
    & 14B & open weight & vLLM & \textbf{\textcolor{OliveGreen}{\cmark}}
    & \nolinkurl{mistralai/Ministral-3-14B-Reasoning-2512} \\
    
GLM-4.6V
    & 106B & open weight & vLLM & 
    & \nolinkurl{zai-org/GLM-4.6V} \\
    
LLaVA-PLLuM-12B
    & 12B & open weight & vLLM & 
    & \nolinkurl{NASK-PIB/LLaVA-PLLuM-12b-nc-instruct-250715} \\

LLaVA-Bielik-11B-v2.6
    & 11B & open weight & vLLM & 
    & \nolinkurl{NASK-PIB/LLaVA-Bielik-11b-v2.6-instruct} \\

InternVL3.5-38B
    & 38B & open weight & vLLM & 
    & \nolinkurl{OpenGVLab/InternVL3_5-38B} \\

\bottomrule
\end{tabularx}

\caption{Specific model versions used in our experiments. The Thinking column indicates models evaluated also in a thinking configuration.}
\label{tab:model-versions}
\end{table*}

\subsection{Random baseline}

We compute the random baseline using the expected accuracy of a uniformly random answer. For yes/no questions, the expected accuracy is \(0.5\). For open-ended questions, it is \(0\), since random generation is not expected to match the reference answer exactly. For a multiple-choice question with \(k\) options (where \(k \in {2,\ldots,8}\)), the expected accuracy is \(1/k\). With circular evaluation, the answer must be correct under all \(k\) cyclic rotations, so the expected accuracy is \((1/k)^k\). We then aggregate these expected per-example accuracies using the same macro-averaging procedure as for model results and report theoretical random accuracy.

\subsection{Prompts and Decoding Parameters}

For open-ended and yes/no examples, for the prompt we provide the raw question:

\begin{quote}
\ttfamily
\{question\}
\end{quote}

For multiple-choice examples, the answer options are appended below the question with letter labels:

\begin{quote}
\ttfamily
\{question\}\\
A. \{option\_A\}\\
B. \{option\_B\}\\
C. \{option\_C\}\\
D. \{option\_D\}\\
...
\end{quote}

Circular evaluation changes only the order and labels of the options.

Images are provided as separate multimodal inputs. Qwen and LLaVA-based vLLM configurations prepend \texttt{<image>} and a newline to the text prompt before applying the chat template:

\begin{quote}
\ttfamily
<image>\\
\{prompt\}
\end{quote}

Other open-weight configurations use the model processor chat template without this prefix. For Qwen and Gemma variants, we set \texttt{enable\_thinking=true} for thinking runs and \texttt{enable\_thinking=false} for non-thinking runs in \texttt{chat\_template\_kwargs}. No few-shot examples or culture-specific hints are added. For our experiments, we use greedy decoding. Model specific decoding parameters are provided in Table~\ref{tab:decoding-parameters}.

\begin{table}[tbh]
\centering
\footnotesize
\setlength{\tabcolsep}{3pt}
\renewcommand{\arraystretch}{1.15}

\begin{tabularx}{\columnwidth}{@{}
    >{\raggedright\arraybackslash}X
    >{\centering\arraybackslash}p{0.16\columnwidth}
    >{\centering\arraybackslash}p{0.11\columnwidth}
    >{\centering\arraybackslash}p{0.11\columnwidth}
@{}}
\toprule
Model & Max tok. & Temp. & Top-$p$ \\
\midrule
GPT-5.4                         & 64 & -- & -- \\
Claude Sonnet 5                 & 64 & -- & -- \\
Qwen3.5-9B                      & 64 & 0.0 & 1.0 \\
Qwen3.5-9B (\textit{Thinking})  & 16384 & 0.0 & 1.0 \\
Qwen3.5-27B                     & 64 & 0.0 & 1.0 \\
Qwen3.5-27B (\textit{Thinking}) & 16384 & 0.0 & 1.0 \\
Qwen3.5-397B-A17B               & 64 & 0.0 & 1.0 \\
Qwen3.5-397B-A17B (\textit{Thinking}) & 16384 & 0.0 & 1.0 \\
Gemma-4-31B-it                  & 64 & 0.0 & 1.0 \\
Gemma-4-31B-it (\textit{Thinking}) & 16384 & 0.0 & 1.0 \\
Mistral-Medium-3.5-128B         & 64 & 0.0 & 1.0 \\
Ministral-3-14B                 & 64 & 0.0 & 1.0 \\
Ministral-3-14B (\textit{Thinking}) & 16384 & 0.0 & 1.0 \\
GLM-4.6V                        & 16384 & 0.0 & 1.0 \\
LLaVA-PLLuM-12B & 64 & 0.0 & 1.0 \\
LLaVA-Bielik-11B-v2.6 & 64 & 0.0 & 1.0 \\
InternVL3.5-38B & 64 & 0.0 & 1.0 \\
\bottomrule
\end{tabularx}

\caption{Decoding parameters used in our experiments. A temperature of 0.0 corresponds to greedy decoding.}
\label{tab:decoding-parameters}
\end{table}

\subsection{Computational Details}

Open-weight models were evaluated using NVIDIA GH200 GPUs with 96 GB of memory. The number of GPUs used in each experiment depended on the size and memory requirements of the model. As a rough estimate of computational cost, a full evaluation run required around 1 GPU-hour for models up to 12B parameters, around 2 GPU-hours for models up to 38B parameters, and around 3--4 GPU-hours for larger models. Reasoning variants were more computationally expensive and required around 8 GPU-hours per run. For proprietary models, we estimate our cost around \$150.

\section{Open-ended Task Error Analysis}

\begin{SaveVerbatim}{JudgePrompt}
You’re a helpful assistant specialized in error analysis.
You will be given:
- a question,
- accepted gold answers,
- an LLM answer.
Your task is to compare the LLM answer with the accepted gold
answers and identify all applicable error types.
The task is multi-label classification: an answer may have more
than one error type. Assign all labels that apply.

Use the following labels:

1) Hallucination
The answer provided by the LLM contradicts the gold answer, contains
factually incorrect information, or gives an entity/value that is not
among the accepted answers.
Example:
Question: Jak nazywa się król przedstawiony na banknocie widocznym
na obrazku?
Gold answer: Zygmunt I Stary
LLM answer: Bolesław Krzywousty
Label: Hallucination

2) Non-adherence to instruction
The answer provided by the LLM can be considered factually correct,
but it does not follow specific instructions included in the prompt
(e.g. incorrect format, too many words, wrong grammatical form, not
answering exactly as requested, answering using whole sentences
instead of one phrase or word, etc.).
Example:
Question: Do jakiej dynastii należał król przedstawiony na banknocie?
Odpowiedz jednym słowem w mianowniku.
Gold answer: Jagiellonowie
LLM answer: Król Zygmunt I Stary należał do dynastii Jagiellonów.
Label: Non-adherence to instruction

3) Language-switching
The answer provided by the LLM can be considered factually correct,
but it is not in Polish.
Example:
Question: Do jakiej dynastii należał król przedstawiony na banknocie?
Odpowiedz jednym słowem w mianowniku.
Gold answer: Jagiellonowie
LLM answer: Jagiellonian dynasty
Label: Language-switching

4) Misspellings
The LLM answer is factually correct, but it contains spelling or
orthographic errors, such as incorrect capitalization or missing
diacritics, and is therefore considered incorrect.
Example:
Question: Jak nazywa się król przedstawiony na banknocie widocznym
na obrazku?
Gold answer: Zygmunt I Stary
LLM answer: Zygmunt I stary
Label: Misspellings

5) Rejection
The LLM answer refuses to answer, states that it cannot identify/
recognize the person, object, or place, says there is not enough
information, or otherwise avoids providing the requested answer.
Example:
Question: Jak nazywa się osoba przedstawiona na zdjęciu? Podaj tylko
imię i nazwisko tej osoby, nic poza tym.
Gold answer: Jan Kowalski
LLM answer: Nie jestem w stanie zidentyfikować osób na podstawie zdjęć.
Label: Rejection

Classification rules:
- Assign Hallucination whenever the answer is factually incorrect or
  contradicts the gold answer.
- Assign Non-adherence to instruction whenever the answer violates
  formatting or instruction requirements specified in the question.
- Assign Language-switching whenever the answer, or a substantial part
  of it, is not in Polish.
- Assign Misspellings whenever the answer is malformed or misspelled.
- Assign Rejection whenever the answer refuses to answer or says it
  cannot identify someone/something.
- Multiple labels may be assigned to the same answer.
- Return valid JSON only, using this schema:
  {"labels": ["..."], "rationale": "one short sentence"}

Question: {question}
Accepted gold answers: {accepted_answers}
LLM answer: {prediction}
\end{SaveVerbatim}

\begin{table}
\centering

\setlength{\fboxsep}{3pt}
\setlength{\fboxrule}{0.3pt}

\fcolorbox{black!25}{black!2}{%
  \begin{minipage}{0.94\columnwidth}
  \UseVerbatim[
    formatcom=\fontsize{5.2pt}{5.6pt}\selectfont,
    breaklines=true,
    breakanywhere=true
  ]{JudgePrompt}
  \end{minipage}%
}

\caption{Prompt used for classifying open-ended answer errors with the judge model.}
\label{tab:open-error-judge-prompt}
\end{table}

\label{app:open-error-analysis}Open-ended questions constitute the most challenging task type in \ourDataset{}, as they require models to generate the target answer directly, in the required grammatical form. To keep the primary benchmark scoring deterministic and reproducible, we constrain the expected output format in the prompt, including the required answer length and, where relevant, grammatical form. We then use deterministic exact-match evaluation against the reference answer, requiring correct Polish diacritics in all cases and correct capitalization where it is part of the expected answer.

We additionally inspect open-ended answers to check whether the scoring mechanism behaves as intended and to understand what types of mistakes models make. We therefore conduct a post-hoc analysis of incorrect or partially invalid open-ended answers. This analysis is not used for benchmark scoring; it is intended only to characterize model behavior and inspect the evaluation protocol.

First, we manually inspected samples of model responses and derived a compact set of recurring error classes:
\begin{itemize}[noitemsep,leftmargin=1em]
    \item \textsc{Hallucination}: the answer introduces an incorrect entity, event, place, object, or cultural association.
    \item \textsc{Non-adherence to instruction}: the output does not follow the required format, for example by providing explanations, multiple candidates, overly long answers, or a different grammatical form than requested.
    \item \textsc{Language switch}: the answer is produced partly or fully outside Polish.
    \item \textsc{Misspelling}: the answer contains orthographic errors, including missing or incorrect diacritics.
    \item \textsc{Rejection}: the model refuses to answer, states that it cannot answer, or claims that the answer cannot be determined despite the task requiring a direct answer.
    \item \textsc{No error}: the model answers correctly and in the required form.
\end{itemize}
Since a single answer may contain more than one problem, the error classes are treated as multi-label classifications. We then apply an LLM-as-a-Judge method using \texttt{DeepSeek-V4-Pro}~\footnote{\url{https://huggingface.co/deepseek-ai/DeepSeek-V4-Pro}}, asking the judge model to assign one or more of the above classes to each open-ended prediction. The prompt used for the judge model is shown in Table~\ref{tab:open-error-judge-prompt}.

\newcommand{\errorcell}[1]{%
  \begingroup
  \pgfmathsetmacro{\errorpct}{%
    max(0,min(100,#1))%
  }%
  \edef\errorcolor{red!\errorpct!white}%
  \expandafter\cellcolor\expandafter{\errorcolor}
  #1%
  \endgroup
}

\begin{table*}
\centering
\scriptsize
\setlength{\tabcolsep}{2.5pt}
\renewcommand{\arraystretch}{1.08}
\resizebox{\textwidth}{!}{%
\begin{tabular}{@{}p{4.2cm}r!{\hspace{2pt}\vrule width 0.35pt\hspace{2pt}}rrrrr@{}}
\toprule
\textbf{Model} & \makecell{\textsc{No}\\\textsc{error}} & \textsc{Halluc.} & \makecell{\textsc{Non-adh.}\\\textsc{instr.}} & \makecell{\textsc{Lang.}\\\textsc{switch}} & \textsc{Misspell.} & \textsc{Reject.} \\
\midrule
Qwen3.5-397B-A17B (\textit{Thinking}) & 64.51 & \errorcell{29.19} & \errorcell{8.46} & \errorcell{1.16} & \errorcell{2.82} & \errorcell{0.00} \\
Gemma-4-31B-it (\textit{Thinking}) & 55.72 & \errorcell{31.67} & \errorcell{14.43} & \errorcell{1.00} & \errorcell{1.49} & \errorcell{7.79} \\
GPT-5.4 & 56.55 & \errorcell{33.00} & \errorcell{8.96} & \errorcell{0.33} & \errorcell{3.65} & \errorcell{2.82} \\
Claude Sonnet 5 & 54.89 & \errorcell{26.70} & \errorcell{18.24} & \errorcell{0.50} & \errorcell{2.82} & \errorcell{7.30} \\
Gemma-4-31B-it & 42.45 & \errorcell{50.25} & \errorcell{13.43} & \errorcell{1.82} & \errorcell{1.16} & \errorcell{1.00} \\
Qwen3.5-397B-A17B & 51.91 & \errorcell{41.63} & \errorcell{11.61} & \errorcell{0.33} & \errorcell{2.32} & \errorcell{0.17} \\
GLM-4.6V & 48.92 & \errorcell{42.29} & \errorcell{14.76} & \errorcell{3.15} & \errorcell{4.64} & \errorcell{2.99} \\
Qwen3.5-27B (\textit{Thinking}) & 42.95 & \errorcell{39.47} & \errorcell{28.36} & \errorcell{9.45} & \errorcell{2.82} & \errorcell{11.94} \\
Qwen3.5-27B & 33.17 & \errorcell{58.37} & \errorcell{15.42} & \errorcell{1.16} & \errorcell{2.65} & \errorcell{0.00} \\
Qwen3.5-9B (\textit{Thinking}) & 31.84 & \errorcell{51.08} & \errorcell{34.83} & \errorcell{13.10} & \errorcell{3.32} & \errorcell{11.94} \\
Qwen3.5-9B & 25.54 & \errorcell{66.83} & \errorcell{16.75} & \errorcell{1.66} & \errorcell{2.49} & \errorcell{0.17} \\
Ministral-3-14B-2512 & 23.55 & \errorcell{67.33} & \errorcell{19.57} & \errorcell{2.16} & \errorcell{4.81} & \errorcell{1.33} \\
LLaVA-Bielik-11B-v2.6 & 22.39 & \errorcell{58.04} & \errorcell{25.54} & \errorcell{2.16} & \errorcell{3.15} & \errorcell{6.47} \\
InternVL3.5-38B & 14.59 & \errorcell{69.65} & \errorcell{22.39} & \errorcell{3.32} & \errorcell{6.97} & \errorcell{4.48} \\
Ministral-3-14B-2512 (\textit{Thinking}) & 20.07 & \errorcell{70.65} & \errorcell{19.73} & \errorcell{5.64} & \errorcell{3.81} & \errorcell{0.50} \\
LLaVA-PLLuM-12B & 15.42 & \errorcell{55.06} & \errorcell{42.79} & \errorcell{1.16} & \errorcell{2.16} & \errorcell{9.95} \\
\midrule
\textbf{Total} & 37.78 & \errorcell{49.45} & \errorcell{19.70} & \errorcell{3.01} & \errorcell{3.19} & \errorcell{4.30} \\
\bottomrule
\end{tabular}
}
\caption{Classified error label frequencies for open-ended predictions on the test split. Values indicate the percentage of predictions assigned each label.}
\label{tab:open-errors}
\end{table*}

The resulting label frequencies are shown in Table~\ref{tab:open-errors}. The results indicate that hallucination is the most frequent error label for open-ended answers. Across all evaluated models, 49.45\% of open-ended predictions are labeled as hallucinated, and the rate exceeds 50\% for several weaker models, including Qwen3.5-9B, Ministral-3-14B-2512, LLaVA-Bielik-11B-v2.6, InternVL3.5-38B, Ministral-3-14B-2512 (\textit{Thinking}), and LLaVA-PLLuM-12B. This suggests that open-ended errors often involve incorrect entities, events, places, objects, or cultural associations.

Instruction non-adherence (19.70\%) is the second most frequent class, appearing more often in lower-performing models, which produce explanations, hedged answers, or responses that do not match the requested concise format. This is visible, for example, for LLaVA-PLLuM-12B (42.79\%), Qwen3.5-9B (\textit{Thinking}) (34.83\%), and Qwen3.5-27B (\textit{Thinking}) (28.36\%). Other errors, such as language switching (3.01\%), misspellings (3.19\%), and rejection (4.30\%), are also observed, but much less often. This suggests that open-ended questions are difficult mainly because models give wrong answers or do not follow the requested format, rather than because of minor spelling or language-form issues.

\section{Results on the Validation Split}
\label{app:validation-split}

\begin{figure}[tbh]
    \centering
    \includegraphics[width=\linewidth]{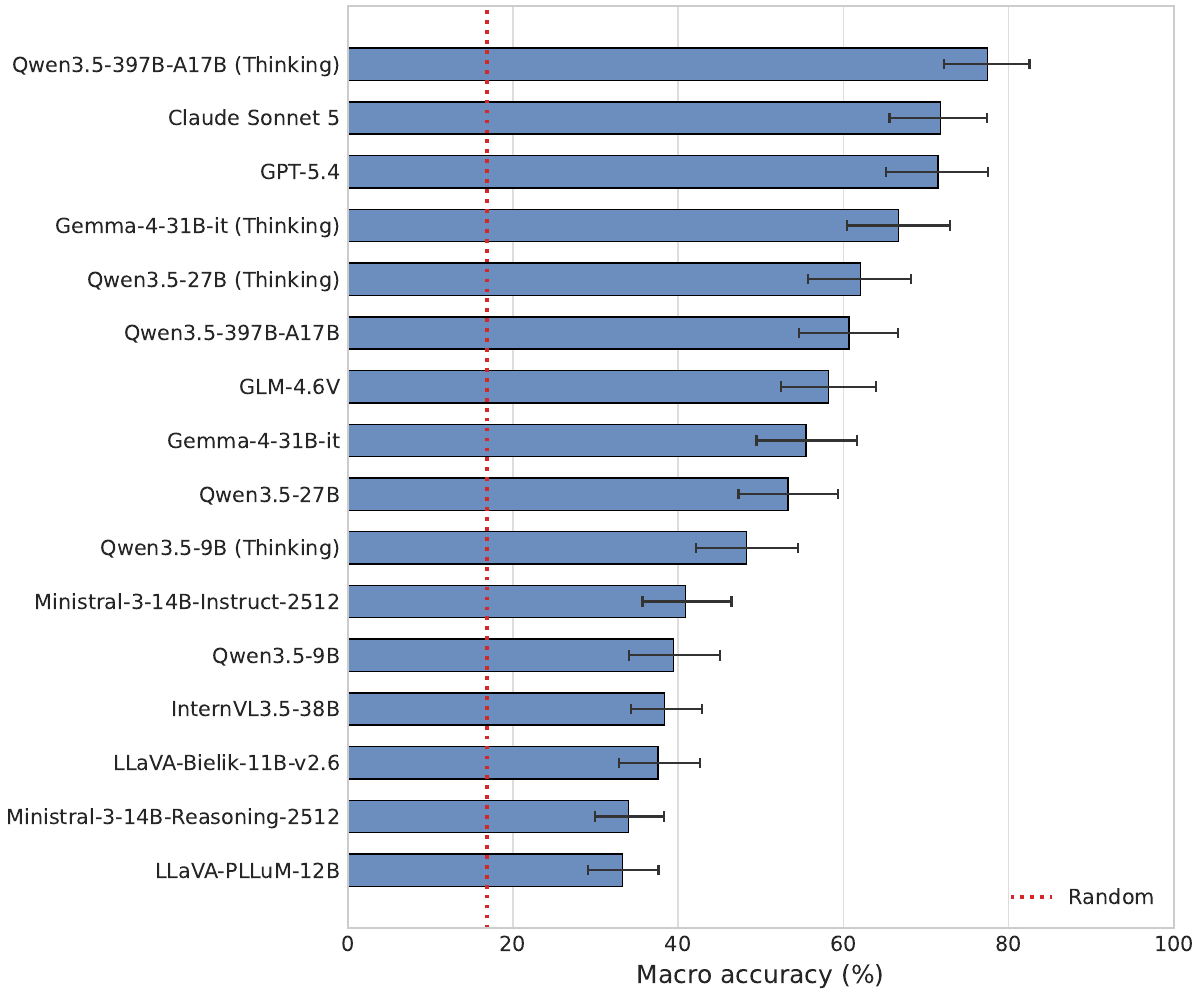}
    \caption{Overall macro accuracy on the validation split with 95\% image-cluster bootstrap confidence intervals. The red dotted line denotes the random baseline.}
    \label{fig:validation-overall-ci-bars}
\end{figure}

This section reports supplementary results for the validation split. As shown in Section~\ref{sec:dataset-statistics}, splitting the validation set by category results in small groups. To obtain more stable estimates, we therefore report confidence intervals only for overall macro accuracy, shown in Figure~\ref{fig:validation-overall-ci-bars}. Table~\ref{tab:validation-results} gives accuracy by dataset category, and Table~\ref{tab:validation-task-results} gives accuracy by task type.

\begin{table*}
\centering
\scriptsize
\setlength{\tabcolsep}{3pt}
\renewcommand{\arraystretch}{1.05}
\resizebox{\linewidth}{!}{
\begin{tabular}{@{}p{5.0cm} C{1.15cm}!{\hspace{2pt}\vrule width 0.35pt\hspace{2pt}} *{7}{C{1.15cm}}@{}}
\toprule
\textbf{Model} & \textbf{Overall} & \makecell{\textbf{Art \&}\\\textbf{Entert.}} & \makecell{\textbf{Culture \&}\\\textbf{Trad.}} & \makecell{\textbf{Geogr. \&}\\\textbf{Nature}} & \makecell{\textbf{History \&}\\\textbf{Society}} & \textbf{Language} & \makecell{\textbf{Image}\\\textbf{Und.}} & \makecell{\textbf{Visual}\\\textbf{Reas.}} \\
\midrule
\midrule
\multicolumn{9}{@{}l}{\textbf{\textsc{Proprietary Models}}} \\
\midrule
Claude Sonnet 5 & \scorecell{71.76} & \scorecell{59.19} & \scorecell{51.94} & \scorecell{91.22} & \scorecell{74.50} & \scorecell{73.39} & \scorecell[\textbf{100.00}]{100.00} & \scorecell{86.67} \\
GPT-5.4 & \scorecell{71.47} & \scorecell[\textbf{62.73}]{62.73} & \scorecell{83.33} & \scorecell[\textbf{94.76}]{94.76} & \scorecell{69.01} & \scorecell{69.12} & \scorecell{86.67} & \scorecell[\textbf{93.33}]{93.33} \\
\midrule
\multicolumn{9}{@{}l}{\textbf{\textsc{Open-Weights Models}}} \\
\midrule
Qwen3.5-397B-A17B (\textit{Thinking}) & \scorecell[\underline{\textbf{77.42}}]{77.42} & \scorecell[\underline{60.22}]{60.22} & \scorecell[\underline{\textbf{86.94}}]{86.94} & \scorecell[\underline{91.98}]{91.98} & \scorecell[\underline{\textbf{88.10}}]{88.10} & \scorecell[\underline{\textbf{77.79}}]{77.79} & \scorecell[\underline{\textbf{100.00}}]{100.00} & \scorecell[\underline{80.00}]{80.00} \\
Gemma-4-31B-it (\textit{Thinking}) & \scorecell{66.70} & \scorecell{54.72} & \scorecell{69.44} & \scorecell{54.35} & \scorecell{71.24} & \scorecell{71.73} & \scorecell{93.33} & \scorecell{63.33} \\
Qwen3.5-27B (\textit{Thinking}) & \scorecell{62.06} & \scorecell{42.07} & \scorecell{66.11} & \scorecell{81.54} & \scorecell{64.94} & \scorecell{66.50} & \scorecell[\underline{\textbf{100.00}}]{100.00} & \scorecell{63.33} \\
Qwen3.5-397B-A17B & \scorecell{60.66} & \scorecell{58.21} & \scorecell{44.72} & \scorecell{54.66} & \scorecell{65.89} & \scorecell{52.06} & \scorecell{86.67} & \scorecell[\underline{80.00}]{80.00} \\
GLM-4.6V & \scorecell{58.18} & \scorecell{54.06} & \scorecell{66.11} & \scorecell{52.96} & \scorecell{66.94} & \scorecell{49.34} & \scorecell{80.00} & \scorecell{63.33} \\
Gemma-4-31B-it & \scorecell{55.49} & \scorecell{40.99} & \scorecell{62.22} & \scorecell{53.27} & \scorecell{53.56} & \scorecell{55.07} & \scorecell{93.33} & \scorecell{76.67} \\
Qwen3.5-27B & \scorecell{53.28} & \scorecell{38.99} & \scorecell{41.11} & \scorecell{82.30} & \scorecell{54.78} & \scorecell{51.72} & \scorecell{93.33} & \scorecell{73.33} \\
Qwen3.5-9B (\textit{Thinking}) & \scorecell{48.24} & \scorecell{35.96} & \scorecell{40.56} & \scorecell{43.59} & \scorecell{60.56} & \scorecell{44.59} & \scorecell{86.67} & \scorecell{56.67} \\
Ministral-3-14B-2512 & \scorecell{40.89} & \scorecell{35.40} & \scorecell{35.00} & \scorecell{77.55} & \scorecell{33.62} & \scorecell{37.16} & \scorecell{86.67} & \scorecell{26.67} \\
Qwen3.5-9B & \scorecell{39.44} & \scorecell{29.66} & \scorecell{29.44} & \scorecell{40.05} & \scorecell{37.18} & \scorecell{36.29} & \scorecell{86.67} & \scorecell{66.67} \\
InternVL3.5-38B & \scorecell{38.36} & \scorecell{36.48} & \scorecell{43.61} & \scorecell{39.11} & \scorecell{24.55} & \scorecell{36.65} & \scorecell{71.67} & \scorecell{43.33} \\
LLaVA-Bielik-11B-v2.6 & \scorecell{37.57} & \scorecell{33.05} & \scorecell{38.89} & \scorecell{45.12} & \scorecell{28.85} & \scorecell{31.04} & \scorecell{71.67} & \scorecell{26.67} \\
Ministral-3-14B-2512 (\textit{Thinking}) & \scorecell{33.99} & \scorecell{28.17} & \scorecell{28.89} & \scorecell{37.14} & \scorecell{34.77} & \scorecell{36.73} & \scorecell{43.33} & \scorecell{33.33} \\
LLaVA-PLLuM-12B & \scorecell{33.24} & \scorecell{32.13} & \scorecell{32.22} & \scorecell{35.57} & \scorecell{29.14} & \scorecell{30.45} & \scorecell{51.67} & \scorecell{13.33} \\
\midrule
Random & \scorecell{16.88} & \scorecell{16.92} & \scorecell{16.83} & \scorecell{16.79} & \scorecell{16.79} & \scorecell{16.97} & \scorecell{16.80} & \scorecell{16.95} \\
\bottomrule
\end{tabular}
}
\caption{Model accuracy by dataset category on the validation split, with all values reported as percentages. The \textbf{best result} in each column is shown in bold, and the best result among open-weight models is \underline{underlined}.}
\label{tab:validation-results}
\end{table*}
\newcommand{\taskcell}[2][\relax]{%
  \begingroup
  \pgfmathsetmacro{\scorepct}{%
    100*max(0,min(1,(#2-50)/50))%
  }%
  \edef\scorecolor{scoreblue!\scorepct!white}%
  \expandafter\cellcolor\expandafter{\scorecolor}
  \ifx#1\relax#2\else#1\fi%
  \endgroup
}

\begin{table*}[htb]
\centering
\scriptsize
\setlength{\tabcolsep}{3pt}
\renewcommand{\arraystretch}{1.05}
\resizebox{\linewidth}{!}{
\begin{tabular}{@{}p{5.0cm} C{1.15cm}!{\hspace{2pt}\vrule width 0.35pt\hspace{2pt}} *{3}{C{1.15cm}}@{}}
\toprule
\textbf{Model} & \textbf{Overall} & \textbf{MCQ} & \textbf{Yes/No} & \textbf{Open} \\
\midrule
\midrule
\multicolumn{5}{@{}l}{\textbf{\textsc{Proprietary Models}}} \\
\midrule
Claude Sonnet 5 & \taskcell{71.76} & \taskcell{67.92} & \taskcell{79.87} & \taskcell{67.50} \\
GPT-5.4 & \taskcell{71.47} & \taskcell[\textbf{70.75}]{70.75} & \taskcell{81.17} & \taskcell{62.50} \\
\midrule
\multicolumn{5}{@{}l}{\textbf{\textsc{Open-Weights Models}}} \\
\midrule
Qwen3.5-397B-A17B (\textit{Thinking}) & \taskcell[\underline{\textbf{77.42}}]{77.42} & \taskcell[\underline{68.40}]{68.40} & \taskcell[\underline{\textbf{86.36}}]{86.36} & \taskcell[\underline{\textbf{77.50}}]{77.50} \\
Gemma-4-31B-it (\textit{Thinking}) & \taskcell{66.70} & \taskcell{66.98} & \taskcell{83.12} & \taskcell{50.00} \\
Qwen3.5-27B (\textit{Thinking}) & \taskcell{62.06} & \taskcell{46.23} & \taskcell{82.47} & \taskcell{57.50} \\
Qwen3.5-397B-A17B & \taskcell{60.66} & \taskcell{49.06} & \taskcell{77.92} & \taskcell{55.00} \\
GLM-4.6V & \taskcell{58.18} & \taskcell{56.60} & \taskcell{77.92} & \taskcell{40.00} \\
Gemma-4-31B-it & \taskcell{55.49} & \taskcell{57.55} & \taskcell{71.43} & \taskcell{37.50} \\
Qwen3.5-27B & \taskcell{53.28} & \taskcell{49.06} & \taskcell{70.78} & \taskcell{40.00} \\
Qwen3.5-9B (\textit{Thinking}) & \taskcell{48.24} & \taskcell{36.79} & \taskcell{77.92} & \taskcell{30.00} \\
Ministral-3-14B-2512 & \taskcell{40.89} & \taskcell{32.55} & \taskcell{70.13} & \taskcell{20.00} \\
Qwen3.5-9B & \taskcell{39.44} & \taskcell{27.83} & \taskcell{62.99} & \taskcell{27.50} \\
InternVL3.5-38B & \taskcell{38.36} & \taskcell{41.98} & \taskcell{65.58} & \taskcell{7.50} \\
LLaVA-Bielik-11B-v2.6 & \taskcell{37.57} & \taskcell{40.57} & \taskcell{57.14} & \taskcell{15.00} \\
Ministral-3-14B-2512 (\textit{Thinking}) & \taskcell{33.99} & \taskcell{25.00} & \taskcell{69.48} & \taskcell{7.50} \\
LLaVA-PLLuM-12B & \taskcell{33.24} & \taskcell{34.43} & \taskcell{57.79} & \taskcell{7.50} \\
\midrule
Random & \taskcell{16.88} & \taskcell{0.65} & \taskcell{50.00} & \taskcell{0.00} \\
\bottomrule
\end{tabular}
}
\caption{Model accuracy by task type on the validation split, with all values reported as percentages. The \textbf{best result} in each column is shown in bold, and the best result among open-weight models is \underline{underlined}.}
\label{tab:validation-task-results}
\end{table*}

\section{Results on the Test Split}
\label{app:detailed_results}

\begin{figure*}
  \centering
  \includegraphics[width=\textwidth]{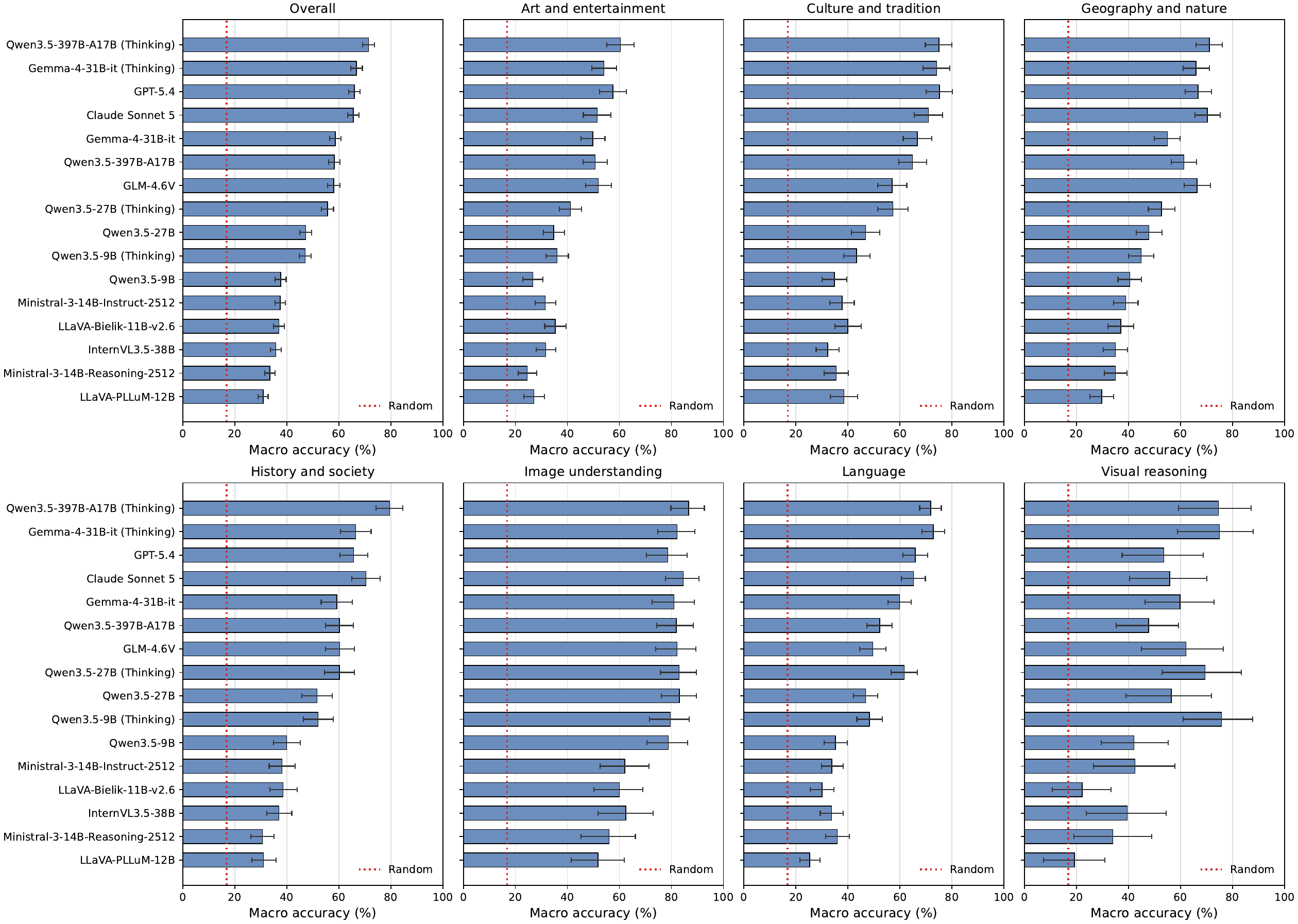}
  \caption{
      Macro accuracy by category with 95\% image-cluster bootstrap confidence intervals on the test split. The red dotted line denotes the random baseline.
  }
  \label{fig:ci-by-category}
\end{figure*}

This section reports supplementary results for the test split. Figure~\ref{fig:ci-by-category} shows category-level macro accuracy with confidence intervals. Table~\ref{tab:test-task-results} gives accuracy by task type.

\begin{table*}
\centering
\scriptsize
\setlength{\tabcolsep}{3pt}
\renewcommand{\arraystretch}{1.05}
\resizebox{\linewidth}{!}{
\begin{tabular}{@{}p{5.0cm} C{1.15cm}!{\hspace{2pt}\vrule width 0.35pt\hspace{2pt}} *{3}{C{1.15cm}}@{}}
\toprule
\textbf{Model} & \textbf{Overall} & \textbf{MCQ} & \textbf{Yes/No} & \textbf{Open} \\
\midrule
\midrule
\multicolumn{5}{@{}l}{\textbf{\textsc{Proprietary Models}}} \\
\midrule
GPT-5.4 & \taskcell{65.93} & \taskcell{62.52} & \taskcell{78.71} & \taskcell{56.55} \\
Claude Sonnet 5 & \taskcell{65.57} & \taskcell{61.28} & \taskcell{80.53} & \taskcell{54.89} \\
\midrule
\multicolumn{5}{@{}l}{\textbf{\textsc{Open-Weights Models}}} \\
\midrule
Qwen3.5-397B-A17B (\textit{Thinking}) & \taskcell[\underline{\textbf{71.45}}]{71.45} & \taskcell[\underline{\textbf{65.94}}]{65.94} & \taskcell[\underline{\textbf{83.89}}]{83.89} & \taskcell[\underline{\textbf{64.51}}]{64.51} \\
Gemma-4-31B-it (\textit{Thinking}) & \taskcell{66.78} & \taskcell{64.07} & \taskcell{80.53} & \taskcell{55.72} \\
Gemma-4-31B-it & \taskcell{58.60} & \taskcell{56.45} & \taskcell{76.89} & \taskcell{42.45} \\
Qwen3.5-397B-A17B & \taskcell{58.25} & \taskcell{45.26} & \taskcell{77.59} & \taskcell{51.91} \\
GLM-4.6V & \taskcell{58.06} & \taskcell{49.77} & \taskcell{75.49} & \taskcell{48.92} \\
Qwen3.5-27B (\textit{Thinking}) & \taskcell{55.65} & \taskcell{45.41} & \taskcell{78.57} & \taskcell{42.95} \\
Qwen3.5-27B & \taskcell{47.19} & \taskcell{43.55} & \taskcell{64.85} & \taskcell{33.17} \\
Qwen3.5-9B (\textit{Thinking}) & \taskcell{47.00} & \taskcell{35.77} & \taskcell{73.39} & \taskcell{31.84} \\
Qwen3.5-9B & \taskcell{37.59} & \taskcell{26.44} & \taskcell{60.78} & \taskcell{25.54} \\
Ministral-3-14B-2512 & \taskcell{37.41} & \taskcell{24.11} & \taskcell{64.57} & \taskcell{23.55} \\
LLaVA-Bielik-11B-v2.6 & \taskcell{36.89} & \taskcell{30.02} & \taskcell{58.26} & \taskcell{22.39} \\
InternVL3.5-38B & \taskcell{35.76} & \taskcell{30.79} & \taskcell{61.90} & \taskcell{14.59} \\
Ministral-3-14B-2512 (\textit{Thinking}) & \taskcell{33.51} & \taskcell{19.13} & \taskcell{61.34} & \taskcell{20.07} \\
LLaVA-PLLuM-12B & \taskcell{30.85} & \taskcell{23.64} & \taskcell{53.50} & \taskcell{15.42} \\
\midrule
Random & \taskcell{16.79} & \taskcell{0.38} & \taskcell{50.00} & \taskcell{0.00} \\
\bottomrule
\end{tabular}
}
\caption{Model accuracy by task type on the test split, with all values reported as percentages. The \textbf{best result} in each column is shown in bold, and the best result among open-weight models is \underline{underlined}.}
\label{tab:test-task-results}
\end{table*}

\section{Detailed Translation Results}
\label{app:translation-results}

Table~\ref{tab:translation-overall-results} reports overall accuracy for the Polish sample and its English and German translations. Deltas are computed relative to the Polish version.

\begin{table*}
\centering
\footnotesize
\resizebox{\textwidth}{!}{%
\begin{tabular}{p{0.4\linewidth}c!{\hspace{4pt}\vrule width 0.3pt\hspace{4pt}}cc!{\hspace{4pt}\vrule width 0.3pt\hspace{4pt}}cc}
\toprule
\textbf{Model} & \textbf{PL}  & \multicolumn{2}{c!{\hspace{4pt}\vrule width 0.3pt\hspace{4pt}}}{\textbf{EN}} & \multicolumn{2}{c}{\textbf{DE}} \\
 &  & \textbf{Value} & \textbf{$\Delta$} & \textbf{Value} & \textbf{$\Delta$} \\
\midrule
Qwen3.5-397B-A17B (\textit{Thinking}) & 75.16 & 68.12 & \textcolor{red!70!black}{-7.04} & 68.15 & \textcolor{red!70!black}{-7.01} \\
Gemma-4-31B-it (\textit{Thinking}) & 72.51 & 72.02 & \textcolor{red!70!black}{-0.48} & 71.55 & \textcolor{red!70!black}{-0.95} \\
Claude Sonnet 5 & 69.94 & 66.94 & \textcolor{red!70!black}{-3.00} & 66.03 & \textcolor{red!70!black}{-3.91} \\
GPT-5.4 & 68.35 & 70.40 & \textcolor{green!50!black}{+2.06} & 63.53 & \textcolor{red!70!black}{-4.82} \\
Gemma-4-31B-it & 63.96 & 60.15 & \textcolor{red!70!black}{-3.82} & 59.04 & \textcolor{red!70!black}{-4.92} \\
Qwen3.5-397B-A17B & 61.89 & 60.56 & \textcolor{red!70!black}{-1.33} & 55.64 & \textcolor{red!70!black}{-6.25} \\
GLM-4.6V & 61.72 & 56.31 & \textcolor{red!70!black}{-5.41} & 58.03 & \textcolor{red!70!black}{-3.69} \\
Qwen3.5-27B (\textit{Thinking}) & 56.65 & 58.35 & \textcolor{green!50!black}{+1.70} & 55.28 & \textcolor{red!70!black}{-1.37} \\
Qwen3.5-27B & 50.83 & 51.84 & \textcolor{green!50!black}{+1.01} & 50.18 & \textcolor{red!70!black}{-0.65} \\
Qwen3.5-9B (\textit{Thinking}) & 49.54 & 51.77 & \textcolor{green!50!black}{+2.23} & 49.44 & \textcolor{red!70!black}{-0.09} \\
LLaVA-Bielik-11B-v2.6 & 40.44 & 39.02 & \textcolor{red!70!black}{-1.42} & 36.94 & \textcolor{red!70!black}{-3.50} \\
InternVL3.5-38B & 39.88 & 41.66 & \textcolor{green!50!black}{+1.79} & 40.81 & \textcolor{green!50!black}{+0.93} \\
Ministral-3-14B-2512 & 39.54 & 41.65 & \textcolor{green!50!black}{+2.11} & 37.66 & \textcolor{red!70!black}{-1.88} \\
Qwen3.5-9B & 38.75 & 47.04 & \textcolor{green!50!black}{+8.29} & 44.83 & \textcolor{green!50!black}{+6.08} \\
Ministral-3-14B-2512 (\textit{Thinking}) & 36.08 & 39.06 & \textcolor{green!50!black}{+2.97} & 37.06 & \textcolor{green!50!black}{+0.98} \\
LLaVA-PLLuM-12B & 34.80 & 32.90 & \textcolor{red!70!black}{-1.90} & 33.08 & \textcolor{red!70!black}{-1.72} \\
\bottomrule
\end{tabular}
}
\caption{Overall model accuracy on the Polish sample and translated English and German samples, with all values reported as percentages. Deltas are computed relative to PL.}
\label{tab:translation-overall-results}
\end{table*}

\end{document}